\documentclass[10pt,twocolumn]{article}
\usepackage{arxiv}
\AtBeginDocument{%
  }

\arxivrunningtitle{SNR-Edit}

\begin{document}

\title{SNR-Edit: Structure-Aware Noise Rectification for Inversion-Free Flow-Based Editing}
\author{%
  Lifan Jiang\textsuperscript{1} \quad
  Boxi Wu\textsuperscript{2,\textdagger} \quad
  Yuhang Pei\textsuperscript{1} \quad
  Tianrun Wu\textsuperscript{2} \\[0.2em]
  Yongyuan Chen\textsuperscript{1} \quad
  Yan Zhao\textsuperscript{3} \quad
  Shiyu Yu\textsuperscript{4} \quad
  Deng Cai\textsuperscript{1}}

\arxivauthornote{\textsuperscript{\textdagger}Corresponding author: Boxi Wu.}
\arxivaffiliations{%
  \begin{tabular}{@{}p{0.47\textwidth}p{0.47\textwidth}@{}}
    \textsuperscript{1}State Key Lab of CAD\&CG, Zhejiang University, Hangzhou, China &
    \textsuperscript{2}School of Software Technology, Zhejiang University, Ningbo, China \\
    \textsuperscript{3}UniTTEC Co., Ltd., Hangzhou, China &
    \textsuperscript{4}Ningbo Meidong Container Terminal Co., Ltd., Ningbo, China
  \end{tabular}}
\arxivcontacts{%
  \begin{tabular}{@{}p{0.47\textwidth}p{0.47\textwidth}@{}}
    \textbf{Lifan Jiang:} \href{mailto:lifanjiang@zju.edu.cn}{lifanjiang@zju.edu.cn};
    \href{https://orcid.org/0009-0003-6090-639X}{ORCID 0009-0003-6090-639X} &
    \textbf{Boxi Wu:} \href{mailto:wuboxi@zju.edu.cn}{wuboxi@zju.edu.cn};
    \href{https://orcid.org/0000-0003-4494-193X}{ORCID 0000-0003-4494-193X} \\
    \textbf{Yuhang Pei:} \href{mailto:3200104179@zju.edu.cn}{3200104179@zju.edu.cn};
    \href{https://orcid.org/0009-0008-9978-5106}{ORCID 0009-0008-9978-5106} &
    \textbf{Tianrun Wu:} \href{mailto:22551034@zju.edu.cn}{22551034@zju.edu.cn};
    \href{https://orcid.org/0009-0005-3834-7362}{ORCID 0009-0005-3834-7362} \\
    \textbf{Yongyuan Chen:} \href{mailto:yongyuan_chen@zju.edu.cn}{yongyuan\_chen@zju.edu.cn};
    \href{https://orcid.org/0009-0007-9778-4284}{ORCID 0009-0007-9778-4284} &
    \textbf{Yan Zhao:} \href{mailto:zhaojing@unittec.com}{zhaojing@unittec.com};
    \href{https://orcid.org/0009-0008-4924-8433}{ORCID 0009-0008-4924-8433} \\
    \textbf{Shiyu Yu:} \href{mailto:15967896108@139.com}{15967896108@139.com};
    \href{https://orcid.org/0009-0004-0039-8538}{ORCID 0009-0004-0039-8538} &
    \textbf{Deng Cai:} \href{mailto:dengcai@cad.zju.edu.cn}{dengcai@cad.zju.edu.cn};
    \href{https://orcid.org/0000-0001-9817-4065}{ORCID 0000-0001-9817-4065}
  \end{tabular}}
\hypersetup{
  pdftitle={SNR-Edit: Structure-Aware Noise Rectification for Inversion-Free Flow-Based Editing},
  pdfauthor={Lifan Jiang, Boxi Wu, Yuhang Pei, Tianrun Wu, Yongyuan Chen, Yan Zhao, Shiyu Yu, Deng Cai}
}

\begin{abstract}
Inversion-free image editing using flow-based generative models offers an efficient alternative to prevailing inversion-based pipelines. However, existing approaches rely on fixed Gaussian noise to initialize the source trajectory. This content-agnostic stochastic noise severely disrupts the spatial topology of the original image, causing severe trajectory drift and structural degradation. To address this source--noise mismatch, we introduce \textbf{SNR-Edit}, a training-free framework that actively prevents structural drift right from the initialization phase. Mechanistically, SNR-Edit performs \textbf{structure-aware noise rectification} by extracting semantic masks and injecting their RoPE-encoded spatial-frequency signatures directly into the initial noise. Instead of applying post-hoc trajectory corrections, this approach provides a stable geometric anchor for the generative flow, explicitly encouraging the model to respect the original layout constraints during source--target transport. This lightweight modification ensures high-fidelity structural preservation without requiring model tuning or inversion. Extensive evaluations across SD3 and FLUX on PIE-Bench and SNR-Bench demonstrate that SNR-Edit achieves superior performance on pixel-level metrics and VLM-based scoring, while adding merely $\sim$1s overhead per image. Our code is released at \url{https://github.com/Tankowa/SNR-Edit}.
\end{abstract}


\begin{teaserfigure}
    \centering
    \includegraphics[width=0.97\textwidth]{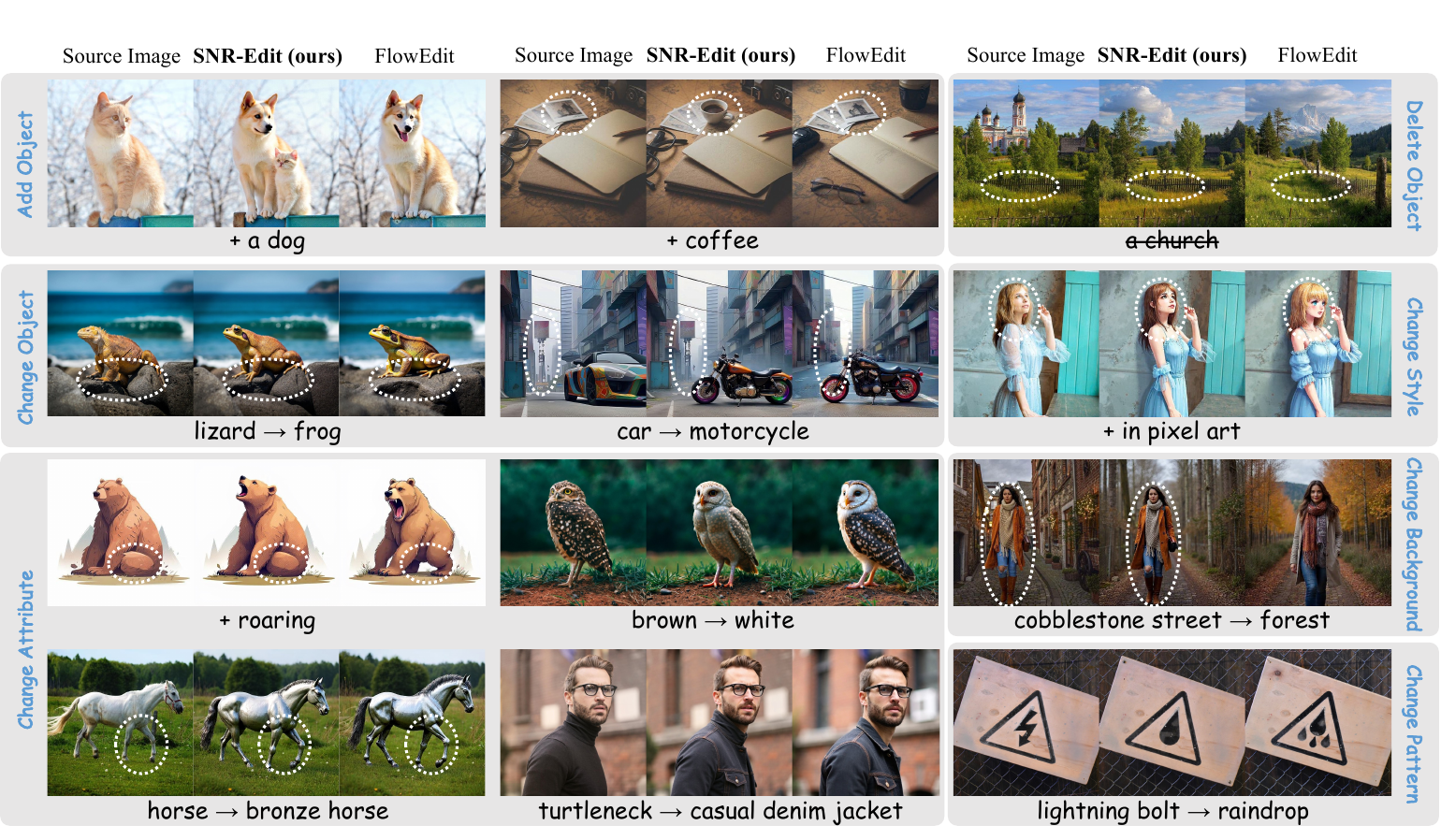}
    \vspace{-5pt}
    \captionof{figure}{\textbf{Qualitative visualization of SNR-Edit results.} Our method demonstrates robust capabilities across a diverse range of tasks, including \textit{Add Object}, \textit{Delete Object}, \textit{Change Object}, \textit{Change Style}, \textit{Change Attribute}, \textit{Change Background}, and \textit{Change Pattern}. Compared to the inversion-free method FlowEdit, SNR-Edit achieves superior structural fidelity, preserving the non-edited layout more faithfully than the baseline while precisely executing the text instructions.}
    \label{fig:teaser}
\end{teaserfigure}

\maketitle

\begin{figure*}[t]
    \centering
    \includegraphics[width=\textwidth]{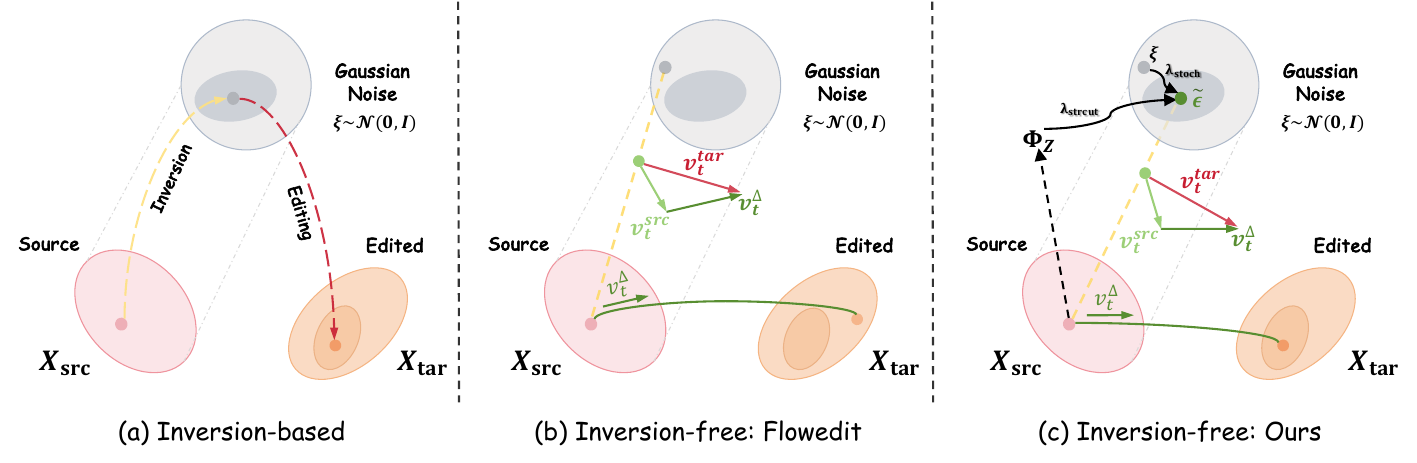}
    \caption{\textbf{Schematic comparison of latent editing dynamics.}
    \textbf{(a) Inversion-based methods} rely on a bidirectional mapping to recover latent noise, yet face an inherent fidelity-editability trade-off and high sensitivity to perturbations, making it difficult to maintain structural consistency.
    \textbf{(b) FlowEdit} circumvents inversion but is hindered by a fundamental \textit{Structural--Stochastic Mismatch}. By initiating the flow from content-agnostic Gaussian noise $\xi \sim \mathcal{N}(0, I)$, the pure stochastic noise destroys the spatial topology of the original image, causing the differential editing trajectory to drift and leading to structural distortion.
    \textbf{(c) Ours.} We introduce structure-aware noise rectification by modulating Gaussian noise with a structural prior $\Phi_{\mathcal{Z}}$ (prepared by resizing, min--max normalization to $[-1,1]$, and channel-wise broadcasting), forming a rectified noise $\tilde{\epsilon} = \lambda_{\text{struct}} \Phi_{\mathcal{Z}} + \lambda_{\text{stoch}} \xi$. This yields a corrected latent source state $\tilde{Z}^{\text{src}}_t = (1 - t) Z_{\text{src}} + t \tilde{\epsilon}$, which provides a stable geometric anchor for flow integration and prevents structural drift right from the initialization phase.}
    \label{fig:method_comparison}
\end{figure*}

\vspace{-5pt}
\section{Introduction}
\label{sec:intro}

With the rapid advancement of flow-based generative models~\cite{flowmatching, rectifiedflow}, such as SD3~\cite{sd3} and FLUX~\cite{flux}, text-guided image editing has advanced toward higher fidelity. The goal is to modify specific semantic attributes of a real image based on textual instructions while preserving the original structural layout and non-edited content. However, existing inversion-free approaches typically initialize the source trajectory with content-agnostic noise, which can weaken structural consistency from the very beginning.

To achieve this goal, existing works have primarily explored distinct technical routes, yet each faces significant limitations in balancing controllability, efficiency, and fidelity.
First, \textit{Inversion-Based Editing} adopts an ``invert-and-generate'' paradigm to map images back to initial noise~\cite{prompt2prompt,rf-inversion,rf-solver}. However, due to the ill-posed nature of inversion, these methods are sensitive to perturbations and suffer from a trade-off between reconstruction fidelity and editability.
Second, methods that integrate \textit{Explicit Structural Conditions}, represented by frameworks such as ControlNet~\cite{controlnet}, offer precise control but incur expensive training overhead and require extensive supervision signals, hindering scalability.
Third, \textit{Architectural Feature Injection}~\cite{pnp} preserves structure without training. However, the deep coupling with specific network architectures makes these strategies difficult to transfer across evolving foundation models.
Finally, recent approaches leveraging \textit{MLLMs}~\cite{instructpix2pix,begal} have improved instruction understanding. Yet, they lack fine-grained spatial perception, resulting in structural inconsistencies such as background drift that fail to meet high-fidelity requirements.

Recently, Inversion-Free Editing has emerged as an alternative. Methods like FlowEdit~\cite{flowedit} construct a mapping path by coupling source and target ODE trajectories without unstable inversion. While concurrent works (e.g., DNAEdit~\cite{dnaedit} and FlowAlign~\cite{flowalign}) mitigate drift from different angles, \textbf{we argue that an under-explored factor is the source--noise mismatch introduced at initialization}. In high-dimensional spaces, adding pure stochastic noise $\xi \sim \mathcal{N}(0, I)$ can disrupt the spatial topology of the original image. Consequently, the integration may start from a structurally weakened state, causing biased trajectory dynamics and structural degradation in non-edited regions. We formalize this limitation at initialization as a \textit{Structural--Stochastic Mismatch}.

To address this, rather than applying post-hoc trajectory corrections, we propose \textbf{SNR-Edit}, a training-free framework that helps prevent structural drift at the source. SNR-Edit achieves \textbf{Latent Trajectory Correction} via segmentation-guided noise control. Instead of adopting passive fixed-noise initialization, we perform structure-aware noise rectification by injecting spatial-frequency constraints (derived from RoPE-encoded semantic masks) directly into the initial noise. This anchors the early-step self-attention mechanisms to the original layout, yielding smoother transport paths and preserving complex geometric structures. As a lightweight plug-in, SNR-Edit retains the efficiency of flow-based editing while improving layout stability and fine-grained fidelity (see~\cref{fig:teaser}).

\begin{figure*}[t]
    \centering
    \includegraphics[width=0.92\textwidth]{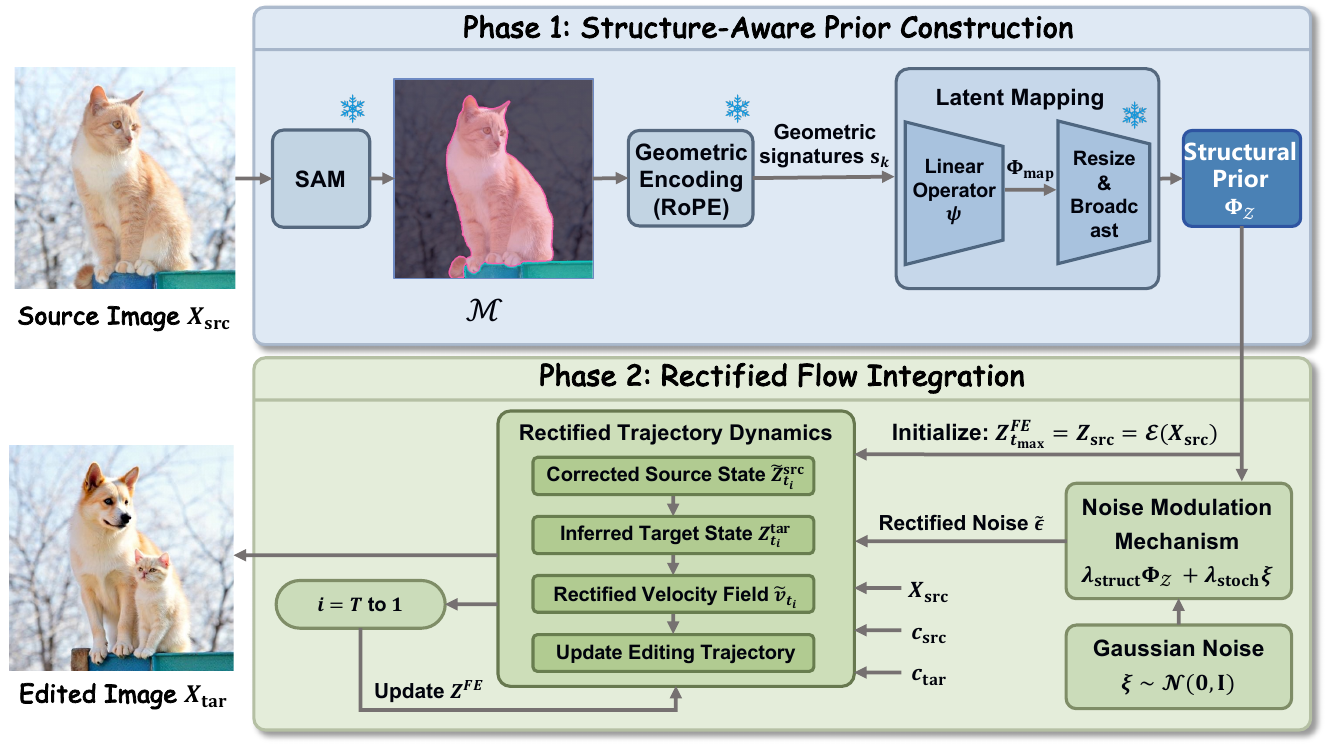}
    \caption{\textbf{SNR-Edit Pipeline (cf.\ Alg.~\ref{alg:nlc}).}
    \textbf{Phase 1} (Top) constructs a structural prior by extracting semantic masks $\mathcal{M}$, computing geometric signatures $\mathbf{s}_k$ via RoPE, and forming a single-channel structural map $\Phi_{\text{map}}$ via a fixed randomized projection $\psi$.
    This map is then resized to the latent resolution, normalized to $[-1,1]$, and broadcast across latent channels to obtain $\Phi_{\mathcal{Z}}$.
    \textbf{Phase 2} (Bottom) executes rectified flow integration in latent space.
    Starting from $Z_{t_{\max}}^{\text{FE}} = Z_{\text{src}}=\mathcal{E}(X_{\text{src}})$, the dynamics iteratively update $Z^{\text{FE}}$ using a rectified velocity field driven by $\tilde{\epsilon}=\lambda_{\text{struct}}\Phi_{\mathcal{Z}}+\lambda_{\text{stoch}}\xi$, which mixes $\Phi_{\mathcal{Z}}$ and Gaussian noise $\xi$.
    This mechanism encourages the output $X_{\text{tar}}=\mathcal{D}(Z^{\text{FE}}_0)$ to faithfully preserve the source layout while realizing the target semantics.}
    \label{fig:pipeline_top}
\end{figure*}

Our contributions can be summarized as follows:
\begin{itemize}
    \item We show that indiscriminate fixed-Gaussian initialization in inversion-free editing can substantially disrupt source spatial topology. We identify this \textit{Structural--Stochastic Mismatch} at initialization as a major contributor to trajectory drift, offering a complementary perspective to existing post-hoc correction methods.
    \item We propose SNR-Edit, a model-agnostic and training-free framework that mitigates structural drift from initialization. By injecting RoPE-encoded semantic priors into the initial noise, it performs adaptive structure-aware noise rectification to anchor the generation process.
    \item We conduct extensive experiments on SD3 and FLUX over PIE-Bench and SNR-Bench. Evaluated by pixel-level metrics, VLM scoring, and a user study, SNR-Edit demonstrates strong overall performance and consistently improves structural preservation with competitive image quality.
\end{itemize}

\begin{figure*}[t]
    \centering
    \includegraphics[width=0.96\textwidth]{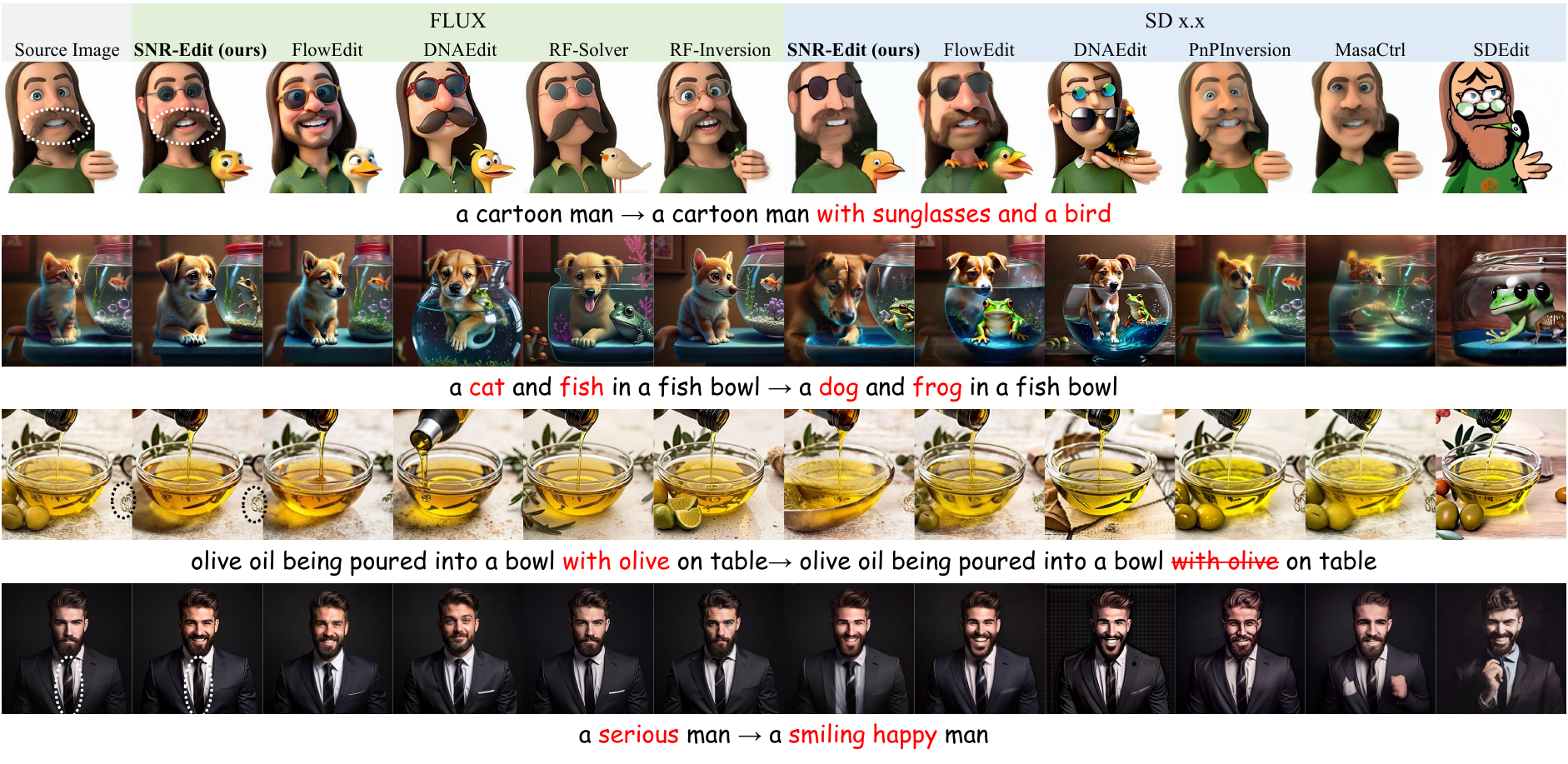}
    \caption{\textbf{Visualization of image editing results on PIE-Bench.} Our method demonstrates superior performance across both FLUX and SD backbones, producing images that better preserve structural details, maintain accurate text-image correspondence, and achieve higher overall visual quality compared to existing approaches.}
    \label{fig:result1}
    \vspace{-3pt}
\end{figure*}

\vspace{-3pt}
\section{Related Work}

\subsection{Inversion-Based Editing Approaches}

Prevailing image editing methods~\cite{null-text,rf-inversion,rf-solver,sdedit,pzero} largely rely on the invert-and-generate paradigm within diffusion models to map images back to their starting noise. Despite optimization techniques designed to reduce reconstruction error, these approaches face an inherent trade-off between fidelity and editability. Crucially, even in the ideal scenario of editing synthetic images where the initial noise is fully known, inversion-based baselines still struggle to maintain structural consistency. This indicates that the standard reverse denoising path is highly sensitive to perturbations, making it difficult to preserve the original layout during text-guided editing.

\subsection{Structure-Aware and MLLM-Guided Approaches}

Current research aiming to enhance editing controllability primarily adopts three strategies, yet each faces significant limitations. First, methods~\cite{controlnet,uni-controlnet,t2iadapter,mou2024t2i,controlnet++} that integrate explicit structural conditions achieve precise control but come at the cost of expensive training overhead, requiring task-specific adapters and extensive supervision signals. Second, other approaches rely on complex architectural designs~\cite{pnp} to inject substantial structural information during the sampling phase. This deep coupling with specific network architectures makes these strategies rigid and difficult to transfer across evolving foundation models. Finally, while approaches leveraging MLLMs~\cite{begal,show-o,qwen-image-edit,gpt4o,editthinker,omnigen2,uniworld} improve the understanding of complex instructions, they typically lack fine-grained spatial perception. This results in poor structural consistency, rendering them unable to consistently meet the strict high-fidelity requirements of precise image editing in highly complex real-world scenarios.

\vspace{-1em}
\subsection{Inversion-Free Flow-Based Editing}

Recently, flow-based generative models~\cite{flowedit,fiaedit,dnaedit} have enabled efficient inversion-free editing by coupling source and target ODE trajectories, thereby circumventing the unstable inversion process. However, early approaches like FlowEdit~\cite{flowedit} rely on a fixed Gaussian noise assumption to approximate the source trajectory. This can disrupt spatial topology and lead to a \textit{Structural--Stochastic Mismatch} (source--noise misalignment), resulting in biased velocity evaluations. Recent advancements such as DNAEdit~\cite{dnaedit} and FlowAlign~\cite{flowalign} mitigate trajectory drift via global velocity guidance or trajectory alignment strategies. Yet these methods act as post-hoc corrections and lack explicit region-aware constraints at initialization. Thus, source--noise misalignment persists, which can impair complex local structures and precise text-image consistency.

\section{Methodology}
\label{sec:method}

In this section, we present SNR-Edit,
a practical framework designed to resolve the conflict between stochastic initialization
and structural preservation in inversion-free editing.
The framework is organized as follows:
We first formalize the limitation of existing inversion-free methods as a
\textbf{Structural--Stochastic Mismatch} in Sec.~\ref{sec:prelim}.
Then, in Sec.~\ref{sec:prior}, we introduce the
\textbf{Structure-Aware Prior Construction} (~\cref{fig:pipeline_top} phase 1),
which injects instance-specific geometric constraints via semantic region decomposition
and randomized geometric projection.
Finally, in Sec.~\ref{sec:dynamics}, we derive the
\textbf{Rectified Flow Integration} (~\cref{fig:pipeline_top} phase 2),
which anchors the editing dynamics in the latent space through structure-aware noise modulation.
We emphasize that SNR-Edit is training-free and does not alter backbone parameters.

\begin{figure}[!t]
    \centering
    \includegraphics[width=0.95\linewidth]{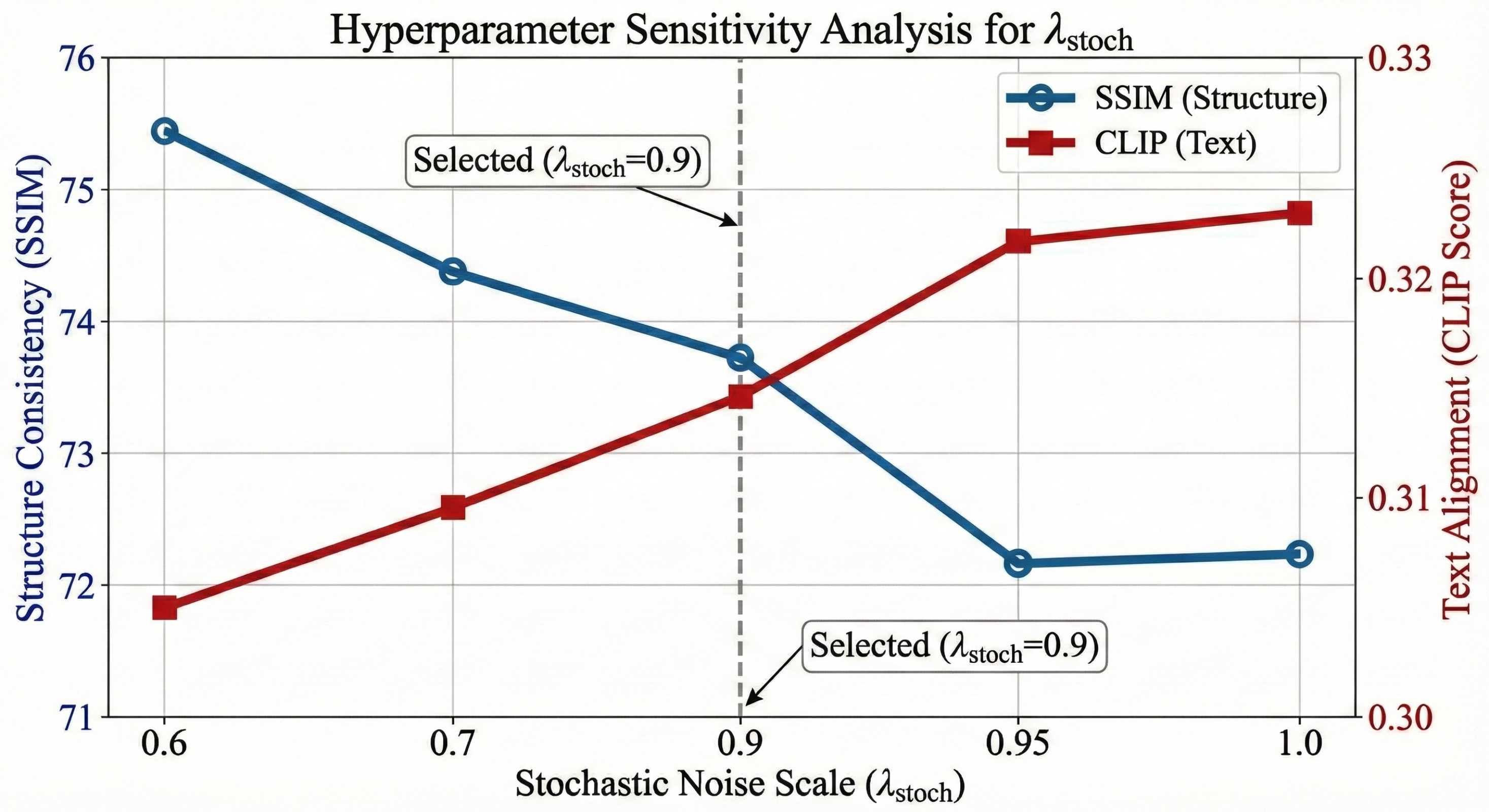}

\caption{\textbf{Sensitivity analysis of the stochastic scale $\lambda_{\text{stoch}}$.} We observe a trade-off between structural consistency (SSIM) and text alignment (CLIP), identifying 0.9 as the optimal balance for text-guided image editing.}
\vspace{-10pt}
    \label{fig:clip}
\end{figure}

\begin{figure*}[t]
    \centering
    \includegraphics[width=0.96\textwidth]{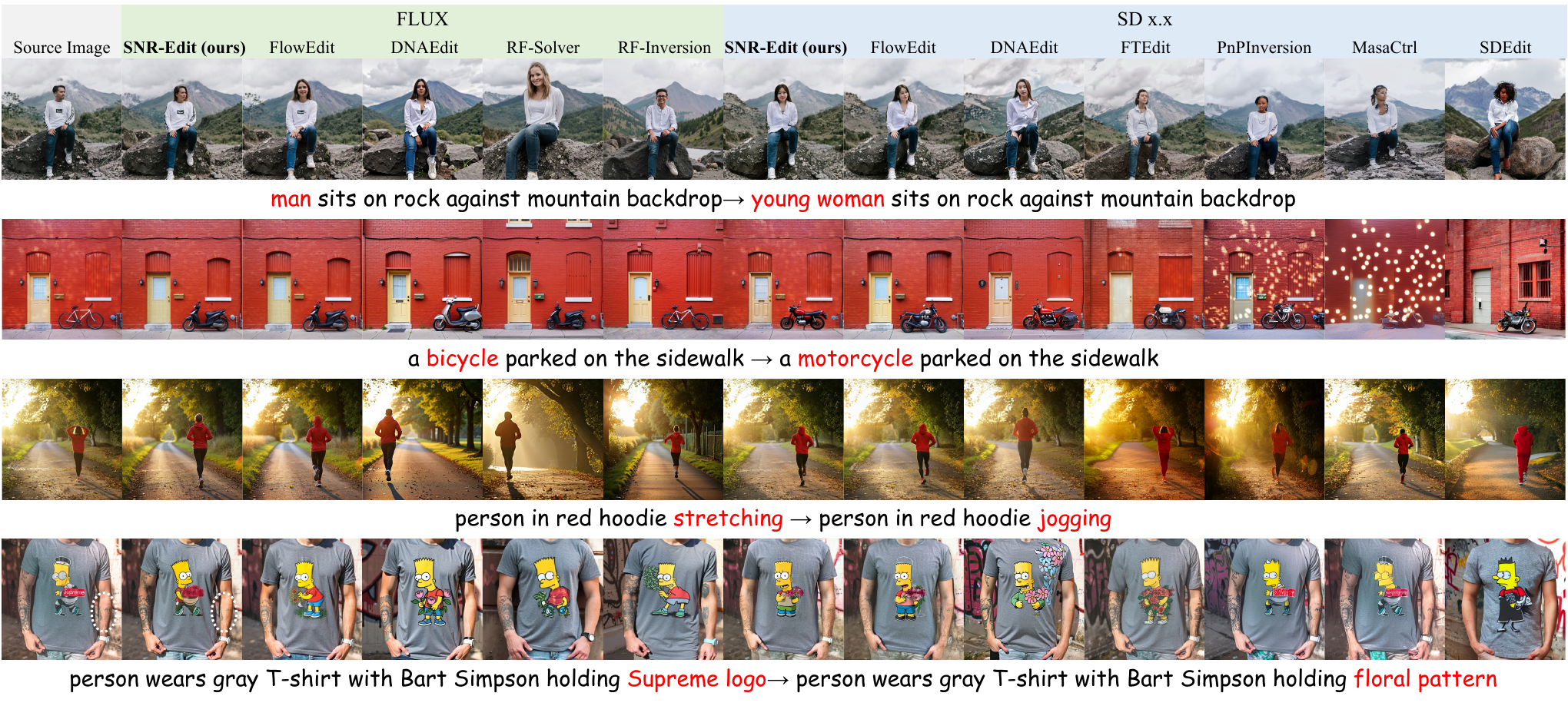}

    \caption{\textbf{Visualization of image editing results on SNR-Bench.} In the above examples, our method demonstrates superior performance on both FLUX and SD architectures, surpassing existing baselines in terms of maintaining structural integrity, aligning with text prompts, and generating high-quality visuals across a diverse set of challenging inputs.}
    \label{fig:result2}

\end{figure*}

\subsection{Preliminaries: Structural--Stochastic Mismatch}
\label{sec:prelim}

Flow Matching models continuously transport a source distribution $p_0$ to a target data distribution $p_1$ 
via a time-dependent velocity field $v(\cdot, t, c)$ conditioned on prompt $c$. 
We operate in the latent space of a pre-trained Autoencoder. Let $Z_{\text{src}} = \mathcal{E}(X_{\text{src}})$ denote the latent representation of the source image $X_{\text{src}}$.
In inversion-free editing, the edited trajectory $Z_t^{\text{FE}}$ is commonly formulated 
using a difference-of-flows ODE dynamic:
\begin{equation}
\label{eq:baseline_ode}
    \dot{Z}_t^{\text{FE}} = 
    v(Z_t^{\text{FE}}, t, c_{\text{tar}}) 
    - v(Z_t^{\text{src}}, t, c_{\text{src}}),
\end{equation}
where $Z_t^{\text{src}}$ denotes a proxy of the source latent trajectory.

Existing inversion-free methods typically approximate the source trajectory by a fixed 
Gaussian interpolation:
\begin{equation}
\label{eq:gaussian_proxy}
    Z_t^{\text{src}} \approx (1 - t) Z_{\text{src}} + t \xi, 
    \quad \xi \sim \mathcal{N}(\mathbf{0}, \mathbf{I}),
\end{equation}
which avoids costly inversion but introduces a fundamental limitation (see ~\cref{fig:method_comparison}(b)).

\vspace{0.2em}
\noindent\textbf{Structural--Stochastic Mismatch.}
Since the Gaussian noise $\xi$ is content-agnostic, the proxy $Z_t^{\text{src}}$ deviates from the local geometric manifold of $Z_{\text{src}}$. As a result, the velocity field $v(\cdot)$ is evaluated off-manifold during integration, leading to trajectory drift and structural degradation (e.g., background distortion) in the edited result. In this paper, we operationalize this off-manifold drift as increased source-to-trajectory discrepancy in early integration steps under frozen prompts. Because exact full-trajectory measurement in high-dimensional latent space is computationally expensive, we provide proxy-based evidence in Appendix \ref{app:drift_discussion}.


\subsection{Structure-Aware Prior Construction}
\label{sec:prior}

To mitigate this mismatch, we propose ~\cref{fig:pipeline_top} phase 1 of our framework:
injecting instance-specific geometric constraints into the stochastic initialization.

\vspace{0.2em}
\noindent\textbf{Semantic Region Decomposition via SAM2.}
We first decompose the source image into stable semantic regions to obtain mask-based geometric anchors.
Given the pixel-space source image $X_{\text{src}} \in \mathbb{R}^{H \times W \times 3}$,
we leverage \textbf{SAM2}~\cite{ravi2024sam2} (specifically the Hiera-Large variant) to explicitly extract instance-level semantic regions.
To ensure robustness, we apply a stability filter (discarding scores $<0.85$) together with simple area-based screening to obtain a filtered set of stable masks $\mathcal{M} = \{ m_1, \dots, m_K \}$.
This mask-based decomposition facilitates reliable boundary delineation in practice, thereby reducing cross-region feature leakage during subsequent processing.

\vspace{0.2em}
\noindent\textbf{Geometric Encoding with RoPE.}
To capture the spatial layout of the source image, we encode the normalized pixel coordinates $\mathbf{p}$ using Rotary Position Embedding (RoPE) into a $C$-dimensional representation. Unlike absolute position encodings, RoPE intrinsically preserves relative spatial distances and directional relationships. For each semantic mask $m_k$, we compute a regional geometric descriptor $\mathbf{s}_k$ by aggregating the embeddings of all enclosed pixels:
\begin{equation}
\label{eq:rope_discrete}
    \mathbf{s}_k = \frac{1}{|m_k|} \sum_{\mathbf{p} \in m_k} \text{RoPE}(\mathbf{p}).
\end{equation}
Functionally, this aggregation transforms discrete pixel coordinates into a continuous spatial-frequency signature that preserves relative layout cues. Rather than merely acting as a bounding box, this descriptor explicitly encodes the region's structural boundaries and internal layout. We empirically validate its utility via ablations (\textbf{Table~\ref{tab:ablation_components}}), where substituting it with naive binary masks results in a noticeable drop in structural fidelity.

\begin{figure}[t]
    \centering
    \includegraphics[width=\linewidth]{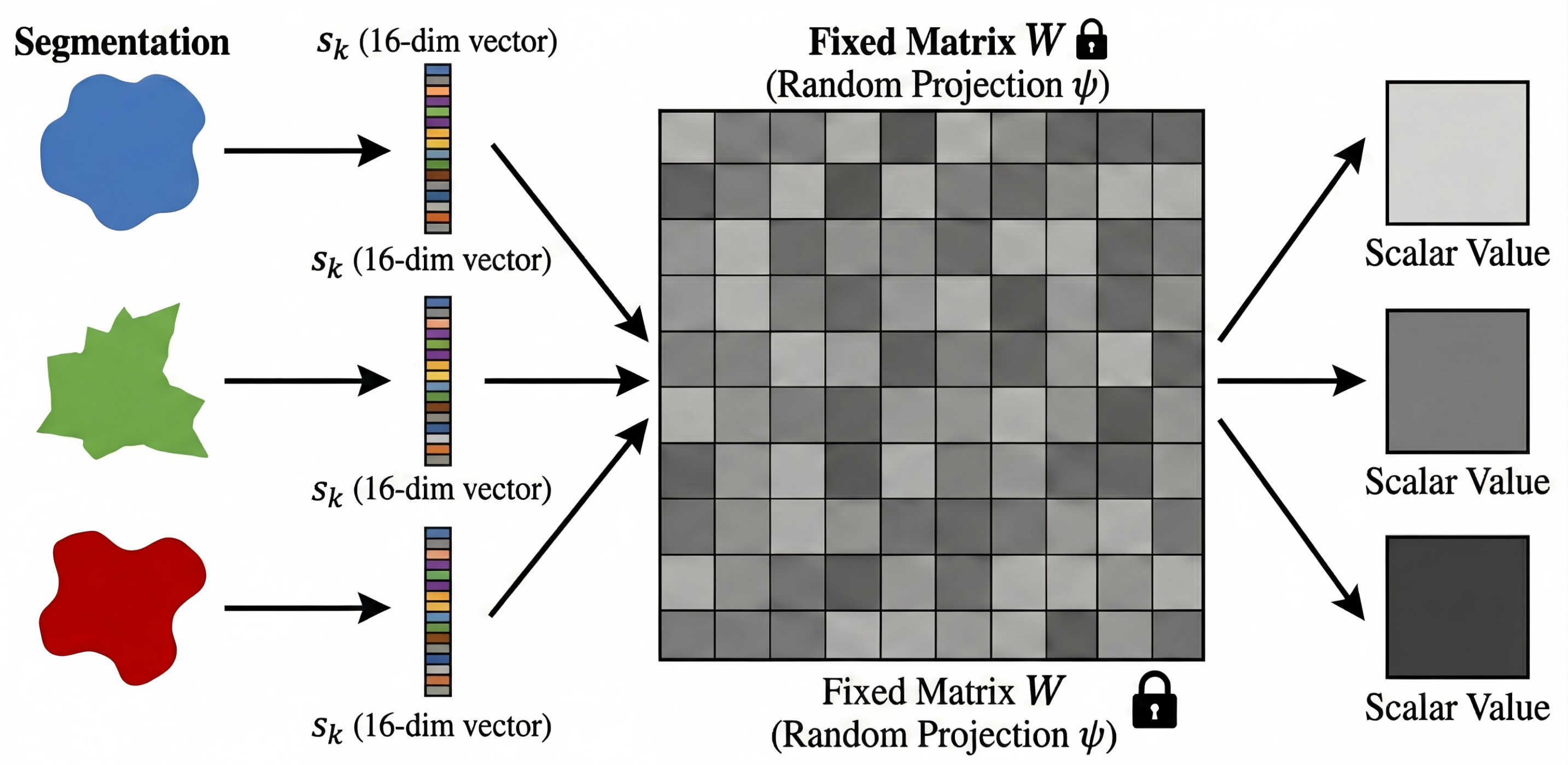}

    \caption{\textbf{Illustration of the Randomized Geometric Projection.} The fixed projection $\psi$, initialized with frozen random weights, maps high-dimensional geometric signatures ($\mathbf{s}_k$) to distinct scalar intensities. This acts as a structural hash, assigning unique identifiers to spatial regions without requiring training or additional supervision.}

    \label{fig:projection_mechanism}
\end{figure}

\vspace{0.2em}
\noindent\textbf{Randomized Geometric Projection.}
To map the high-dimensional geometric signatures into a latent structural constraint, we project each $\mathbf{s}_k$ into a scalar intensity using a fixed, randomly initialized linear weight vector $W \in \mathbb{R}^{1 \times C}$ drawn from $\mathcal{U}(-\frac{1}{\sqrt{C}}, \frac{1}{\sqrt{C}})$. This assigns a distinct scalar value (in practice) to each semantic region without requiring training.

In practice, SAM2 may occasionally generate overlapping masks for adjacent objects (e.g., a single pixel might be assigned to both a ``person'' and a ``chair'' mask). To prevent the blending of distinct structural features, we resolve these overlaps by assigning the shared pixels exclusively to the mask with the highest SAM confidence score. This strict assignment ensures that every pixel is deterministically mapped to a single, distinct structural intensity. We discuss projected-region ID separability and empirical collision behavior under our mask-filtered setting in Appendix \ref{app:collision_rate}, where region confusion is shown to be rare in practice. The resulting non-overlapping structural map is formulated as:
\begin{equation}
\label{eq:phi_map}
    \Phi_{\text{map}}(\mathbf{p}) =
    \sum_{k=1}^{K}
    \psi(\mathbf{s}_k)\,
    \mathbb{I}_{m_k}(\mathbf{p}).
\end{equation}
Finally, we spatially resize $\Phi_{\text{map}}$ to the latent resolution, normalize it to the $[-1,1]$ interval via min--max scaling (with a zero-map fallback when the range is negligible), and broadcast it across latent channels to obtain the latent structural prior $\Phi_{\mathcal{Z}}$, strictly matching the exact shape of the latent noise.

\begin{algorithm}[!h]
   \caption{Inversion-Free Editing via SNR-Edit}
   \label{alg:nlc}
\begin{algorithmic}[1]
    \linespread{1.05}\selectfont
   \STATE {\bfseries Input:} Source Image $X_{\text{src}}$, Prompts $c_{\text{src}}, c_{\text{tar}}$, Steps $\{t_i\}_{i=0}^T$, Scales $\lambda_{\text{struct}}, \lambda_{\text{stoch}}$
   \STATE {\bfseries Output:} Edited image $X_{\text{tar}}$
   \STATE {\bfseries Init:} Encode source $Z_{\text{src}} \leftarrow \mathcal{E}(X_{\text{src}})$
   \STATE {\bfseries Init:} $Z_{t_{\max}}^{\text{FE}} \leftarrow Z_{\text{src}}$ \COMMENT{Initialize trajectory}
   
   \STATE \textcolor{blue}{\textsc{// Phase 1: Structure-Aware Prior Construction}}
   \STATE $\mathcal{M} \leftarrow \text{Segment}(X_{\text{src}})$ \COMMENT{SAM2 on pixel space}
   \STATE $\Phi_{\mathcal{Z}} \leftarrow \text{ComputePrior}(\mathcal{M}, \text{RoPE})$ \COMMENT{Obtain latent prior via \cref{eq:phi_map}}

   \STATE \textcolor{blue}{\textsc{// Phase 2: Rectified Flow Integration}}
   \FOR{$i = T$ {\bfseries to} $1$}
       \STATE $\xi \sim \mathcal{N}(\mathbf{0}, \mathbf{I})$
       \STATE $\tilde{\epsilon} \leftarrow \lambda_{\text{struct}} \Phi_{\mathcal{Z}} + \lambda_{\text{stoch}} \xi$ \COMMENT{Noise Modulation}
       
       \STATE $\tilde{Z}_{t_i}^{\text{src}} \leftarrow (1 - t_i)Z_{\text{src}} + t_i \tilde{\epsilon}$ \COMMENT{Corrected latent source state}
       
       \STATE \textcolor{gray}{\textsc{// Calculate Structural Offset}}
       \STATE $\Delta \tilde{Z}_{t_i} \leftarrow \tilde{Z}_{t_i}^{\text{src}} - Z_{\text{src}}$ \COMMENT{Dynamic geometric anchor}
       
       \STATE \textcolor{gray}{\textsc{// Rectified Velocity Evaluation}}
       \STATE $\tilde{v}_{t_i} \leftarrow v(Z_{t_i}^{\text{FE}} + \Delta \tilde{Z}_{t_i}, t_i, c_{\text{tar}}) - v(\tilde{Z}_{t_i}^{\text{src}}, t_i, c_{\text{src}})$
       
       \STATE $Z_{t_{i-1}}^{\text{FE}} \leftarrow Z_{t_i}^{\text{FE}} + (t_{i-1} - t_i)\tilde{v}_{t_i}$ \COMMENT{Update latent trajectory}
   \ENDFOR
   
   \STATE {\bfseries Return:} $X_{\text{tar}} = \mathcal{D}(Z_0^{\text{FE}})$ \COMMENT{Decode to pixel space}
\end{algorithmic}
\end{algorithm}

\begin{table*}[!t]
\centering

\renewcommand{\arraystretch}{0.98} 
\caption{\textbf{Quantitative comparison on PIE-Bench.} We evaluate methods across SD 1.x, SD3, and FLUX backbones. Methods are categorized by their mechanism: \textbf{Inv-B} (Inversion-Based), \textbf{Inv-F} (Inversion-Free). \textbf{Bold} indicates the best result, and \underline{underline} indicates the second best across all methods (global comparison). The \textbf{Avg. Rank} is calculated across all methods (global ranking). Overall, \textbf{SNR-Edit} shows strong structural-preservation performance with competitive alignment quality.}
\label{tab:quantitative_pie}

\resizebox{\textwidth}{!}{%
\begin{tabular}{c|l|c|cccccc|c}
\toprule
\textbf{Backbone} & \textbf{Method} & \textbf{Type} & \textbf{Dist} ($\times 10^3$) $\downarrow$ & \textbf{PSNR} $\uparrow$ & \textbf{LPIPS} ($\times 10^3$) $\downarrow$ & \textbf{MSE} ($\times 10^4$) $\downarrow$ & \textbf{SSIM} ($\times 10^2$) $\uparrow$ & \textbf{CLIP} $\uparrow$ & \textbf{Avg. Rank} $\downarrow$ \\
\midrule
\multirow{3}{*}{\textbf{SD 1.x}} 
& SDEdit (SD1.5) & Inv-B & 197.01 & 14.43 & 538.76 & 2754.35 & 43.41 & 0.314 & 9.83 \\
& MasaCtrl (SD1.4) & Inv-B & 119.37 & 18.85 & 267.56 & 1011.02 & 68.98 & 0.293 & 5.83 \\
& PnPInversion (SD1.4) & Inv-B & \underline{93.93} & \underline{20.93} & \underline{211.94} & \underline{627.28} & 73.13 & 0.304 & \underline{3.33} \\
\midrule
\midrule
\multirow{3}{*}[-2pt]{\textbf{SD3}} 
& DNAEdit & Inv-F & 164.61 & 15.96 & 452.57 & 1871.58 & 58.82 & \underline{0.318} & 8.50 \\
& FlowEdit & Inv-F & 128.69 & 18.17 & 300.39 & 1162.54 & 72.23 & \textbf{0.323} & 5.67 \\
\cmidrule{2-10}
& \textbf{SNR-Edit (Ours)} & Inv-F & 121.85 & 18.62 & 279.12 & 1037.96 & \underline{73.74} & 0.315 & 4.50 \\
\midrule
\midrule
\multirow{5}{*}[-2pt]{\textbf{FLUX}} 
& RF-Inversion & Inv-B & 113.69 & 19.36 & 342.16 & 929.17 & 57.61 & 0.288 & 6.67 \\
& RF-Solver & Inv-B & 108.80 & 19.77 & 352.47 & 853.59 & 70.62 & 0.305 & 5.00 \\
& DNAEdit & Inv-F & 158.02 & 16.38 & 435.90 & 1742.18 & 61.90 & 0.313 & 8.17 \\
& FlowEdit & Inv-F & 129.82 & 18.19 & 299.42  & 1199.97 & 73.33 & 0.307 & 6.17 \\
\cmidrule{2-10}
& \textbf{SNR-Edit (Ours)} & Inv-F & \textbf{91.35} & \textbf{21.32} & \textbf{195.36} & \textbf{607.07} & \textbf{79.96} & 0.294 & \textbf{2.33} \\
\bottomrule
\end{tabular}%
}
\end{table*}

\subsection{Rectified Flow Integration}
\label{sec:dynamics}

In ~\cref{fig:pipeline_top} phase 2, we perform editing by rectifying the stochastic initialization used in the inversion-free dynamics.

\vspace{0.2em}
\noindent\textbf{Structure-Aware Noise Rectification.}
During prior preparation, we resize $\Phi_{\mathcal{Z}}$ to the latent resolution and normalize it to the bounded range $[-1, 1]$ via min--max scaling (with a zero-map fallback when the value range is negligible). This bounded preprocessing keeps the structural signal at a stable magnitude comparable to latent noise used by the pre-trained model, thereby mitigating out-of-distribution behavior (see analysis in Appendix \ref{app:theoretical_analysis}).
We use the rectified noise to build a corrected source state before evaluating the flow dynamics. We then construct the rectified noise by combining this normalized prior
with Gaussian noise (see~\cref{fig:method_comparison}(c); analysis in~\cref{fig:clip}):

\begin{equation}
\label{eq:rectified_noise}
    \tilde{\epsilon} =
    \lambda_{\text{struct}} \, \Phi_{\mathcal{Z}}
    + \lambda_{\text{stoch}} \, \xi,
    \quad \xi \sim \mathcal{N}(\mathbf{0}, \mathbf{I}).
\end{equation}

\vspace{0.2em}
\noindent\textbf{Intuition: Why does structure-aware noise preserve layout?}
In Diffusion Transformers (DiTs), the initial self-attention maps are sensitive to the spatial distribution of the input noise. When relying solely on pure Gaussian noise, the model lacks explicit spatial guidance during early denoising steps and often shows structural drift. Injecting region-specific, RoPE-encoded intensities into $\tilde{\epsilon}$ introduces a spatial bias that helps query and key tokens recognize object boundaries and coarse layout from early steps.

Using this rectified noise $\tilde{\epsilon}$, we compute the corrected source state $\tilde{Z}_t^{\text{src}}$ following the update rule in \textbf{Alg.~\ref{alg:nlc}}, which serves as a stable geometric anchor for subsequent flow integration.

\vspace{0.2em}
\noindent\textbf{Rectified Editing Dynamics.} 
To ensure strict structural alignment, we introduce a rectified flow dynamic that compensates for geometric deviations. Specifically, we evaluate the target velocity not at the drifting current state, but at a corrected position ``re-anchored'' to the source trajectory. This forces the generative flow to respect the original layout constraints while evolving semantic content. We reformulate the editing ODE as:
\textit{Implementation:} We discretize Eq.~\eqref{eq:snr_ode} over $\{t_i\}$ and re-sample $\xi$ each step (Alg.~\ref{alg:nlc}).
\begin{equation}
\label{eq:snr_ode}
    \dot{Z}_t^{\text{FE}} = 
    v(Z_t^{\text{FE}} + \Delta \tilde{Z}_t, t, c_{\text{tar}}) 
    - v(\tilde{Z}_t^{\text{src}}, t, c_{\text{src}}),
\end{equation}

where $\Delta \tilde{Z}_t$ represents the time-dependent structural offset in the latent space, defined as:
\begin{equation}
\label{eq:delta_z}
    \Delta \tilde{Z}_t = \tilde{Z}_{t}^{\text{src}} - Z_{\text{src}}.
\end{equation}
Instead of relying on ambiguous definitions of latent manifolds, this formulation provides a concrete physical mechanism: it re-anchors the target velocity evaluation directly to the corrected source trajectory. By explicitly compensating for the geometric deviation $\Delta \tilde{Z}_t$ at each step, SNR-Edit strongly encourages the generative flow to more closely respect the original layout constraints while evolving the semantic content towards the target prompt. A Lipschitz stability analysis further indicates that the re-anchored evaluation in Eq.~\eqref{eq:snr_ode} yields a controlled vector-field error and a bounded trajectory deviation (Appendix \ref{app:reanchoring_bound}, Eq.~\eqref{eq:field_error_bound_app}--\eqref{eq:traj_error_bound_app}).

\section{Experiments}
\label{sec:experiments}

\subsection{Experimental Settings}
\label{sec:settings}

\textbf{Datasets and Benchmarks.}
We evaluate on two benchmarks spanning different complexity levels. First, we use \textbf{PIE-Bench}~\cite{pie} (700 images) for standard quantitative evaluation. Second, to address the limited resolution and semantic diversity of existing sets, we construct \textbf{SNR-Bench}. Rather than prioritizing sheer scale, SNR-Bench serves as a compact yet highly representative stress test of 80 high-quality images: $\sim$50\% sampled from PIE-Bench for continuity, and $\sim$50\% web-collected to encompass broader semantic and structural diversity. We cover four editing operations: \textit{adjust}, \textit{change}, \textit{remove}, and \textit{add}. Prompts for non-PIE images are manually verified to eliminate noise and reduce ambiguity.

\begin{figure}[t]  
    \centering
    \includegraphics[width=\linewidth]{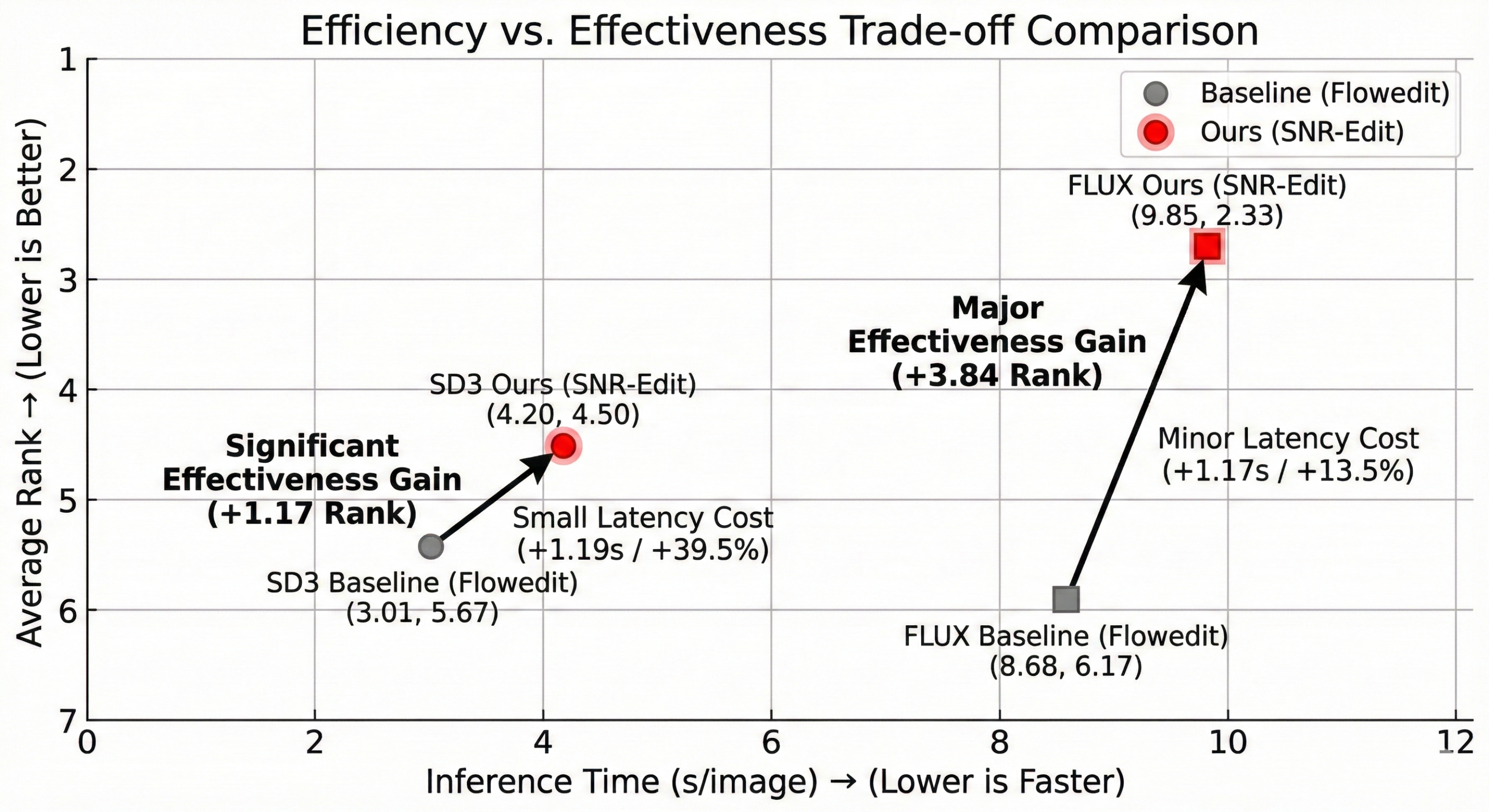}
    \caption{\textbf{Efficiency vs. Effectiveness Trade-off on PIE-Bench.}
    The comparisons are conducted on a single A100 GPU. The axes contrast inference latency against the Average Rank (sourced from~\cref{tab:quantitative_pie}). 
    Compared to FlowEdit, SNR-Edit achieves significant performance gains with only a marginal increase in latency. 
    This indicates that our method offers a highly favorable trade-off, significantly enhancing generation quality at an acceptable speed cost on both SD3 and FLUX architectures.}
    \label{fig:sanwei}
\end{figure}

\vspace{0.2em}
\noindent\textbf{Baselines and Implementation.}
We implement \textbf{SNR-Edit} on SD3~\cite{sd3} and FLUX~\cite{flux}. We compare against inversion-based methods (RF-Inversion~\cite{rf-inversion}, RF-Solver~\cite{rf-solver}, SDEdit~\cite{sdedit}, PnPInversion~\cite{pnp}, MasaCtrl~\cite{Masactrl}, FTEdit~\cite{FTEdit}) and inversion-free methods (DNAEdit~\cite{dnaedit}, FlowEdit~\cite{flowedit}). For fairness, inversion-free baselines use unified sampling configurations on the same base model, while inversion-based methods follow their original settings.

\vspace{0.2em}
\noindent\textbf{Ablation Variants.}
To validate each component's effectiveness in our structure-aware prior, we evaluate three variants on the SD3 backbone:
(1) \textit{w/o Semantic Decomp.} removes Semantic Region Decomposition, applying RoPE and projection to the global image;
(2) \textit{w/o RoPE} discards geometric embeddings, relying solely on binary masks from SAM;
(3) \textit{w/o Rand. Proj.} replaces the randomized linear projection with average pooling.

\vspace{0.2em}
\noindent\textbf{Evaluation Metrics.}
We assess structural consistency, background preservation, and semantic alignment using three protocols:
(1) \textit{Quantitative Metrics:} On PIE-Bench, we report latent distance (structure), PSNR/SSIM/MSE/LPIPS (background/perceptual quality), and CLIP similarity (text-image alignment).
(2) \textit{VLM Evaluation:} On SNR-Bench, we use ImgEdit Reward~\cite{imgedit} and Qwen3-VL-32B-Instruct~\cite{qwen3vl} to score Structural Fidelity, Text-Image Alignment, and Background Consistency, which helps capture global coherence not fully reflected by low-level metrics.
(3) \textit{User Study:} We conduct a randomized blind user study with 38 valid participants (from 40 recruited). Each participant completed 20 randomized comparison tasks under a strictly anonymous protocol.

\subsection{Main Results}
\label{sec:results}

\begin{table*}[t]
\renewcommand{\arraystretch}{0.89}
\centering
\caption{\textbf{Perceptual Evaluation via VLMs and User Study.}
We report assessment results (1--5) from two VLM-based evaluators
(ImgEdit Reward and Qwen3-VL-32B-Instruct) and a human user study.
Methods are grouped by backbone: SD x.x, SD3, and FLUX.
\textbf{Bold} indicates the best result, and \underline{underline}
indicates the second best across all methods.
The \textbf{Avg. Rank} is calculated over all nine criteria using
global competition ranking, where tied scores receive the same rank.}
\label{tab:mllm_user}

\resizebox{\textwidth}{!}{%
\begin{tabular}{c|l|ccc|ccc|ccc|c}
\toprule
& & \multicolumn{3}{c|}{\textbf{ImgEdit Reward Model}}
& \multicolumn{3}{c|}{\textbf{Qwen3-VL-32B-Instruct}}
& \multicolumn{3}{c|}{\textbf{User Study}} & \\
\cmidrule(lr){3-5}
\cmidrule(lr){6-8}
\cmidrule(lr){9-11}
\cmidrule(lr){12-12}
\textbf{Backbone} & \textbf{Method}
& \textbf{Struct} $\uparrow$ & \textbf{Text} $\uparrow$ & \textbf{BG} $\uparrow$
& \textbf{Struct} $\uparrow$ & \textbf{Text} $\uparrow$ & \textbf{BG} $\uparrow$
& \textbf{Struct} $\uparrow$ & \textbf{Text} $\uparrow$ & \textbf{Qual} $\uparrow$
& \textbf{Avg. Rank} $\downarrow$ \\
\midrule

\multirow{4}{*}{\textbf{SD x.x}}
& SDEdit (SD1.5)
& 2.53 & 2.49 & 2.69 & 1.63 & 2.51 & 1.31
& 2.17 & 2.83 & 2.42 & 11.33 \\
& MasaCtrl (SD1.4)
& 3.04 & 2.46 & 3.31 & 2.49 & 1.95 & 2.59
& 2.68 & 2.11 & 2.57 & 9.56 \\
& PnPInversion (SD1.4)
& \underline{3.48} & 3.03 & 3.80 & \underline{3.58} & 2.84 & 2.94
& 3.64 & 2.97 & 3.21 & 5.67 \\
& FTEdit (SD3.5)
& \textbf{3.55} & \underline{3.30} & 3.99 & \underline{3.58} & 3.23 & 3.25
& \underline{3.72} & 3.46 & 3.79 & 2.89 \\

\midrule
\midrule

\multirow{3}{*}[-2pt]{\textbf{SD3}}
& DNAEdit
& 2.75 & 2.70 & 2.98 & 2.30 & 3.21 & 1.80
& 2.63 & 3.37 & 2.91 & 8.67 \\
& FlowEdit
& 3.34 & 3.19 & 3.90 & 3.35 & \textbf{3.65} & 3.28
& 3.48 & \textbf{3.82} & 3.54 & 4.11 \\
\cmidrule{2-12}
& \textbf{SNR-Edit (Ours)}
& 3.44 & \underline{3.30} & \underline{4.08}
& 3.40 & \underline{3.58} & \underline{3.30}
& 3.69 & \underline{3.73} & \underline{3.87} & \underline{2.56} \\

\midrule
\midrule

\multirow{5}{*}[-2pt]{\textbf{FLUX}}
& RF-Inversion
& 3.36 & 2.94 & 3.70 & 2.99 & 2.63 & 2.31
& 3.26 & 2.94 & 3.12 & 7.67 \\
& RF-Solver
& 2.42 & 2.40 & 2.47 & 2.41 & 3.19 & 1.89
& 2.51 & 3.14 & 2.68 & 10.00 \\
& DNAEdit
& 2.86 & 2.85 & 3.16 & 2.03 & 3.20 & 1.75
& 2.92 & 3.19 & 2.86 & 8.67 \\
& FlowEdit
& 3.38 & 3.21 & 4.04 & 3.28 & 3.39 & 3.19
& 3.53 & 3.58 & 3.41 & 4.33 \\
\cmidrule{2-12}
& \textbf{SNR-Edit (Ours)}
& \textbf{3.55} & \textbf{3.38} & \textbf{4.24}
& \textbf{3.80} & 3.09 & \textbf{3.80}
& \textbf{3.94} & 3.42 & \textbf{3.98} & \textbf{2.22} \\
\bottomrule
\end{tabular}%
}
\end{table*}

\textbf{Visualization of image editing results.}
\cref{fig:result1,fig:result2} show representative edits on PIE-Bench and SNR-Bench for SD3 and FLUX. SNR-Edit better preserves identity, geometry, and high-frequency details while implementing the requested semantic changes. In contrast, baselines more often exhibit identity drift, local distortion, or over-smoothing in fine textures. The efficiency analysis in \cref{fig:sanwei} further indicates that these qualitative gains are achieved with only a small latency increase over FlowEdit, offering a favorable quality--efficiency trade-off.

\begin{table}[t]
    \centering
    \renewcommand{\arraystretch}{1.2}
    \caption{\textbf{Ablation Study.} We analyze the contribution of each module in the structure-aware prior construction phase using the SD3 backbone. The ``Baseline'' represents the method without any proposed components (equivalent to FlowEdit). Text-Image alignment scores are evaluated by Qwen3VL-32B to measure semantic alignment.}

    \label{tab:ablation_components}
    \resizebox{1.0\linewidth}{!}{
    \begin{tabular}{l|cccc}
        \toprule
        \textbf{Method Variant} & \textbf{Dist} ($\times 10^3$) $\downarrow$ & \textbf{SSIM} ($\times 10^2$) $\uparrow$ & \textbf{LPIPS} ($\times 10^3$) $\downarrow$ & \textbf{Text} $\uparrow$ \\
        \midrule
        \textbf{Baseline (FlowEdit)} & 128.69 & 72.23 & 300.39 & 3.386 \\
        \midrule
        \textbf{w/o Semantic Decomp.} & 130.47 & 71.93 & 303.63 & 3.309 \\
        \textbf{w/o RoPE} (Binary Masks) & 129.12 & 72.67 & 290.21 & 3.210 \\
        \textbf{w/o Rand. Proj.} (Avg Pooling) & 128.82 & 72.58 & 291.27 & 3.251 \\
        \midrule
        \rowcolor{gray!10} \textbf{SNR-Edit (Ours)} & \textbf{121.85} & \textbf{73.74} & \textbf{279.12} & \textbf{3.568} \\
        \bottomrule
    \end{tabular}
    }

\end{table}

\vspace{0.2em}
\noindent\textbf{Quantitative Analysis on Pixel-Level Metrics.}
Table~\ref{tab:quantitative_pie} shows that SNR-Edit leads inversion-free baselines on PIE-Bench. Compared with FlowEdit on FLUX, it reduces Structural Distance by $29.6\%$ and LPIPS by $34.8\%$, improving the average rank from $6.17$ to $2.33$. The gains extend beyond distance-based metrics: PSNR increases from $18.19$ to $21.32$, SSIM rises from $73.33$ to $79.96$, and MSE decreases from $1199.97$ to $607.07$. The same trend holds on SD3, where Structural Distance, LPIPS, and MSE decrease by $5.3\%$, $7.1\%$, and $10.7\%$, respectively. Moreover, SD3-based SNR-Edit improves PSNR from $18.17$ to $18.62$ and SSIM from $72.23$ to $73.74$, demonstrating that the preservation gains generalize across both backbones. Although CLIP is slightly lower on both backbones, these trends show that the gains concentrate on structural preservation and support the intended preservation--alignment trade-off.

\vspace{0.2em}
\noindent\textbf{Perceptual Evaluation with VLM Reward Models.}
Table~\ref{tab:mllm_user} reports the SNR-Bench results obtained using ImgEdit Reward~\cite{imgedit} and Qwen3-VL-32B-Instruct~\cite{qwen3vl}. With FLUX, SNR-Edit improves over FlowEdit by $0.17$ and $0.20$ on ImgEdit structure and background, and by $0.52$ and $0.61$ on the corresponding Qwen3-VL scores. On SD3, its ImgEdit scores also improve by $0.10$, $0.11$, and $0.18$ in structure, text alignment, and background consistency, respectively. On FLUX, the ImgEdit text score increases from $3.21$ to $3.38$, whereas the Qwen3-VL text score decreases from $3.39$ to $3.09$, indicating evaluator-dependent variation in semantic assessment. Overall, the VLM results show that SNR-Edit provides more consistent improvements in structural fidelity and background preservation, while its semantic-alignment performance varies across evaluators.

\vspace{0.2em}
\noindent\textbf{User Study Validation.}
We conducted a randomized, blinded user study with 38 participants, each completing 20 comparison tasks under an anonymous protocol. As shown in Table~\ref{tab:mllm_user}, SNR-Edit receives the highest structural and quality scores in both the SD3 and FLUX groups. On FLUX, its structural score rises from $3.53$ to $3.94$ and quality from $3.41$ to $3.98$ relative to FlowEdit. Although FlowEdit retains higher text-alignment preference, these results confirm stronger structural preservation and overall visual quality.

\vspace{0.2em}
\noindent\subsection{Ablation Study}
\label{sec:ablation}

Table~\ref{tab:ablation_components} quantifies the contribution of each module on the SD3 backbone. Removing Semantic Decomposition causes the largest degradation (Dist $130.47$ and LPIPS $303.63$), supporting the need for region-level priors. Replacing RoPE with binary masks lowers SSIM to $72.67$ and Text to $3.210$, showing that binary boundaries do not fully capture relative spatial relationships. Replacing randomized projection with average pooling yields Dist $128.82$ and Text $3.251$, confirming that average pooling weakens region discrimination. The text-alignment score drops in all ablations. Compared with FlowEdit, the full design reduces Dist by $5.3\%$ and LPIPS by $7.1\%$. Notably, the complete model achieves the highest SSIM ($73.74$) and Text score ($3.568$), further validating the joint design across complementary evaluation criteria. Together, these results show that the components are complementary, and full SNR-Edit best balances structural fidelity, perceptual quality, and semantic alignment.

\section{Conclusion}
\label{sec:conclusion}

We identify the Structural--Stochastic Mismatch induced by the fixed Gaussian proxy in inversion-free editing as a key bottleneck that biases transport dynamics and leads to structural drift. To address this, we propose \textbf{SNR-Edit}, a lightweight, training-free, and model-agnostic framework that anchors inversion-free editing with an instance-specific geometric prior. Our method constructs a spatially guided latent prior via semantic region decomposition and spatial encoding, and rectifies stochastic initialization through fixed-ratio noise modulation. By replacing the biased source proxy with a structure-aware trajectory anchor, SNR-Edit stabilizes the editing dynamics and improves structural consistency without the computational overhead of inversion or fine-tuning. Extensive experiments on SD3 and FLUX, evaluated via pixel-level metrics, VLM-based scoring, and a user study, show gains in structural preservation and perceptual quality, with mixed semantic alignment. These results highlight the value of trajectory engineering via structure-aware initialization for robust, efficient, and controllable flow-based image editing. Its lightweight design also preserves the efficiency advantages that make inversion-free editing attractive in practical applications. In practice, SNR-Edit is most effective when segmentation quality is reliable and the target edits do not require large-scale geometric recomposition.

\vspace{1.0em}
\noindent\textbf{Limitations and Societal Impact.}
While SNR-Edit mitigates structural drift, its performance depends on the quality of the semantic masks produced by SAM2. Under extreme occlusions, transparent surfaces, or highly complex scenes, inaccurate segmentation may introduce noisy structural priors and reduce the method to behavior closer to conventional inversion-free editing. Representative failure cases include inaccurate object boundaries caused by imperfect masks, residual changes in background contrast, and occasional checkerboard artifacts. Addressing these limitations remains an important direction for improving robustness in challenging real-world scenes. Finally, the ability to perform high-fidelity structural editing raises concerns about malicious image manipulation and deceptive content, underscoring the need for reliable AIGC detection tools and responsible disclosure practices.


\bibliographystyle{ACM-Reference-Format}
\bibliography{arxiv}


\appendix

\section{SNR-Bench Case Showcase}
\label{app:snrbench_cases_en}

\textbf{SNR-Bench} comprises 80 high-quality image-editing cases. Approximately 50\% are sampled from PIE-Bench to ensure continuity with standard benchmarks, and the remaining 50\% are collected from the web to introduce richer textures and more complex real-world scenes. We cover four editing operations: \textit{adjust}, \textit{change}, \textit{remove}, and \textit{add}. To minimize ambiguity and improve instruction consistency, all editing instructions for the non--PIE-Bench subset are \textbf{manually written, refined, and verified} through human annotation.

\subsection{What we show.}
In this section, we provide representative visual examples selected from the 80 cases. Each example includes the source image $X_{\text{src}}$, the original/edited prompts (with the edited instruction $c_{\text{edit}}$), and edited outputs $X_{\text{edit}}$ from different methods. The showcased cases span diverse edit types and difficulty levels, illustrating the challenges in SNR-Bench and complementing our quantitative results on structural fidelity, text-image alignment, and background consistency.

\subsection{Case showcase.}
\begin{figure*}[t]
    \centering
    \includegraphics[width=1.0\linewidth]{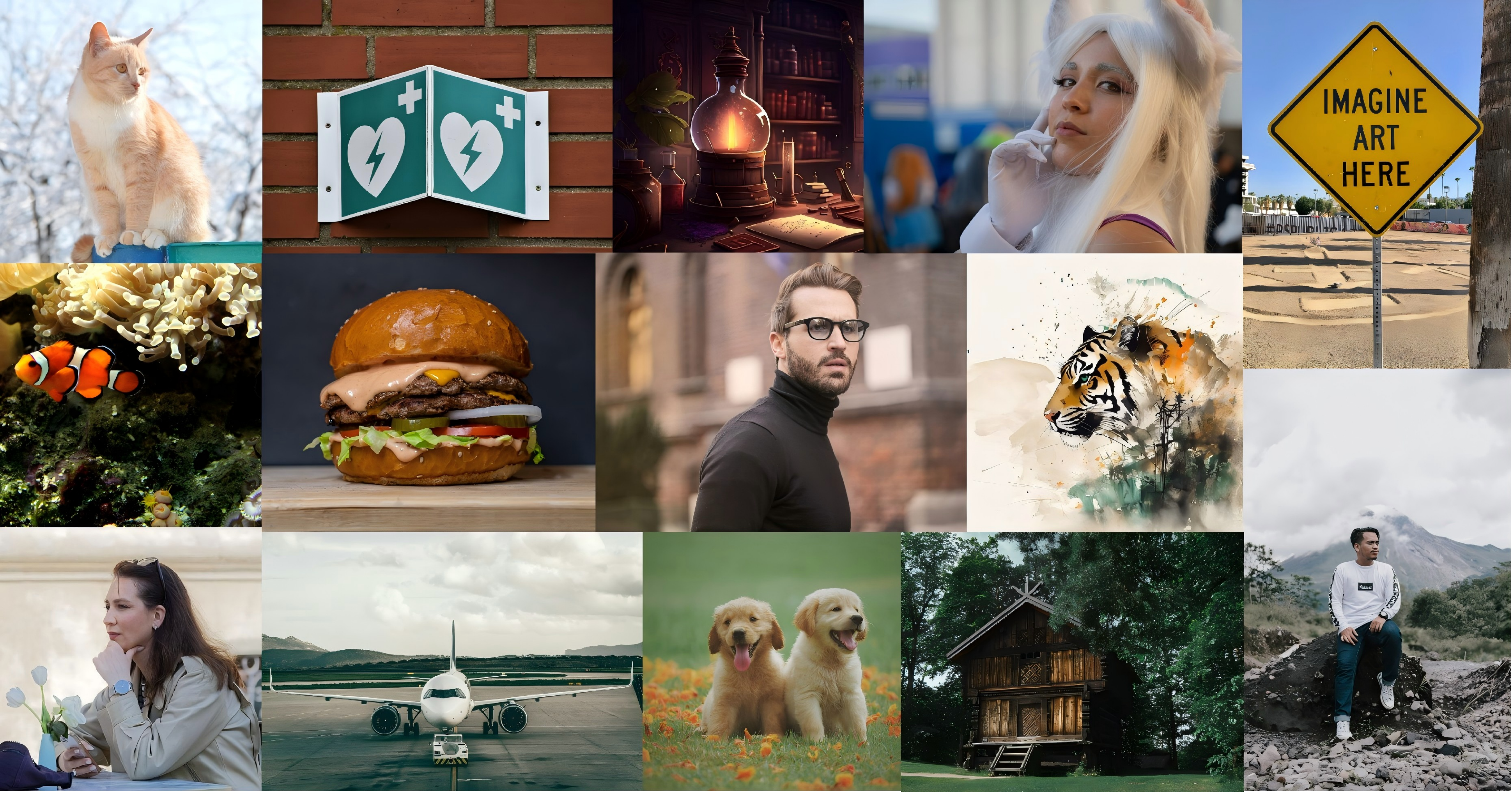}
    \vspace{-0.2cm}
    \caption{\textbf{Overview of SNR-Bench Diversity.} We display representative source images from our constructed benchmark, covering six major categories: \textit{Animals}, \textit{Humans}, \textit{Landscapes}, \textit{Architecture}, \textit{Patterns}, and \textit{OCR/Text}. This diversity ensures a robust evaluation of structural consistency across both organic textures and rigid geometric layouts.}
    \label{fig:snr_bench_overview}
\end{figure*}

To demonstrate the diversity and complexity of our evaluation set, Figure~\ref{fig:snr_bench_overview} provides a gallery of representative source images from SNR-Bench. The dataset is designed to cover a broad spectrum of semantic domains and structural types, ensuring a comprehensive assessment of editing capabilities. As shown in the figure, the cases are categorized into distinct groups:

\begin{itemize}
    \item \textbf{Animals \& Humans:} Images containing complex biological structures that require fine-grained preservation of identity and texture.
    \item \textbf{Landscapes \& Architecture:} Scenes with rich depth information and rigid geometric structures, testing the model's ability to maintain spatial layout.
    \item \textbf{Patterns \& Text (OCR):} Highly structured inputs such as repetitive geometric patterns and legible text. These cases serve as rigorous stress tests for structural fidelity, as even minor trajectory drifts can lead to noticeable distortions or legibility loss.
\end{itemize}

These diverse categories ensure that SNR-Edit is evaluated across varying levels of difficulty, ranging from organic object manipulation to rigid structural preservation.

\section{Theoretical Analysis of Rectified Noise Distribution}
\label{app:theoretical_analysis}

In this section, we provide a theoretical justification for the proposed \textbf{Structure-Aware Noise Rectification (SNR)} mechanism.
A primary concern when injecting structural priors into the initialization of flow-based models is the potential \textit{distribution shift} (OOD issue).
Since pre-trained backbones are optimized under the assumption that the stochastic component is sampled as $\xi \sim \mathcal{N}(0, I)$, substantially altering the input scale may lead to artifacts.
Throughout this appendix, we use $v_\theta(\cdot,t,c)$ interchangeably with $v(\cdot,t,c)$ in the main paper.

We show that our \textbf{prior preparation} (resize + min--max normalization + broadcasting) constrains the structural prior to a bounded range consistent with the latent noise scale used by the backbone, while preserving relative structural semantics across regions.
In addition, we provide a simple Lipschitz stability bound that justifies the \textbf{re-anchored velocity evaluation} in \cref{eq:snr_ode} (cf.~\cref{app:reanchoring_bound}).

\subsection{Structure-Preserving Min--Max Normalization}

Let $\Phi_{\text{map}} \in \mathbb{R}^{H \times W}$ denote the raw single-channel structural map constructed in Phase~1 (cf.~\cref{eq:phi_map}).
A naive standardization approach (e.g., Z-score normalization) would enforce zero mean and unit variance:
\begin{equation}
    \hat{\Phi}_{z\text{-}score} = \frac{\Phi_{\text{map}} - \mu_{\Phi}}{\sigma_{\Phi}}.
\end{equation}
However, such normalization can undesirably amplify low-contrast regions or suppress informative inter-region contrasts, weakening the relative structural cues carried by the region-wise intensities.
Instead, we adopt a \textbf{range-constrained min--max normalization} consistent with our implementation.

Specifically, we first resize $\Phi_{\text{map}}$ to the latent resolution, denoted as $\Phi_{\text{map}}^{\downarrow}$, and normalize it to the bounded interval $[-1,1]$:
\begin{equation}
\begin{aligned}
\Phi_{\text{norm}}
&= 2\,
\frac{
  \Phi_{\text{map}}^{\downarrow}
  - \min(\Phi_{\text{map}}^{\downarrow})
}{
  \max(\Phi_{\text{map}}^{\downarrow})
  - \min(\Phi_{\text{map}}^{\downarrow}) + \varepsilon
}
- 1.
\end{aligned}
\end{equation}
with a zero-map fallback when $\max(\Phi_{\text{map}}^{\downarrow})-\min(\Phi_{\text{map}}^{\downarrow})$ is negligible.
This affine mapping yields two key properties:
\begin{enumerate}
    \item \textbf{Preservation of Relative Structure:} Min--max normalization is order-preserving and applies an affine transform, thus maintaining region-wise intensity ordering and relative contrasts that encode coarse geometry.
    \item \textbf{Bounded Magnitude:} The structural signal is explicitly bounded as $\Phi_{\text{norm}}(x)\in[-1,1]$ for any spatial location $x$, preventing uncontrolled magnitude growth.
\end{enumerate}

Finally, we broadcast $\Phi_{\text{norm}}$ across latent channels to obtain the latent structural prior $\Phi_{\mathcal{Z}}$ used in Phase~2; this operation changes only tensor shape, not value range.
By construction, $\Phi_{\mathcal{Z}}$ is already at the latent resolution after resizing $\Phi_{\text{map}}$ and broadcasting.

\subsection{Defense Against Distribution Shift and Rectification Stability}

We model the rectified noise $\tilde{\epsilon}$ as a linear combination of the bounded structural prior $\Phi_{\mathcal{Z}}$ and the stochastic component $\xi \sim \mathcal{N}(0, I)$:
\begin{equation}
\label{eq:rectified_noise_app}
    \tilde{\epsilon} = \lambda_{\text{struct}} \Phi_{\mathcal{Z}} + \lambda_{\text{stoch}} \xi.
\end{equation}
We then form the corrected source state $\tilde{Z}_t^{\text{src}}=(1-t)Z_{\text{src}}+t\tilde{\epsilon}$ and the offset $\Delta\tilde{Z}_t=\tilde{Z}_t^{\text{src}}-Z_{\text{src}}$, consistent with \cref{eq:snr_ode,eq:delta_z}.

\textbf{Bounded Prior, Controlled Mixing.}
By construction, $\Phi_{\mathcal{Z}}$ is bounded element-wise in $[-1,1]$.
Therefore, the structural term has a controlled amplitude determined by $\lambda_{\text{struct}}$, while the stochastic term remains Gaussian.
This design avoids introducing heavy-tailed or arbitrarily large perturbations that could push latent features into extreme activation regimes of the pre-trained backbone.

\textbf{Energy Consistency.}
The min--max normalization establishes a unified and bounded numerical scale for the structural prior, so the mixing coefficients $\lambda_{\text{stoch}}$ and $\lambda_{\text{struct}}$ operate on consistent magnitudes.
Consequently, the prior is neither overwhelmed by Gaussian noise (if too small) nor dominates to cause artifacts (if too large), promoting stable rectification during ODE integration.

\textbf{Practical Robustness.}
Although the resulting $\tilde{\epsilon}$ is not strictly Gaussian due to the additive bounded component, the rectification acts as a mild, controlled bias on the initialization that preserves geometric cues without inducing severe OOD behavior.
Empirically, this bounded preprocessing is sufficient to maintain stable backbone evaluation while improving structural fidelity.

\subsection{Re-anchored Velocity Evaluation: A Lipschitz Stability Bound}
\label{app:reanchoring_bound}

This subsection provides a citable justification for the \textbf{re-anchored velocity evaluation} in \cref{eq:snr_ode}.
We show that, under a standard Lipschitz assumption on the backbone velocity field, re-anchoring yields a controlled vector-field error and thus a bounded trajectory deviation.

\paragraph{Setup.}
Let the ideal source trajectory be $Z_t^{\text{src}}$, and define the ideal inversion-free editing dynamics
\begin{equation}
\label{eq:ideal_ode_app}
\begin{aligned}
\dot{Z}_t^{\star}
&= v_\theta\!\left(Z_t^{\star},t,c_{\text{tar}}\right) \\
&\quad - v_\theta\!\left(Z_t^{\text{src}},t,c_{\text{src}}\right).
\end{aligned}
\end{equation}
In practice, we only have a proxy source trajectory $\tilde{Z}_t^{\text{src}}$.
Our method uses the \emph{re-anchoring offset}
\begin{equation}
\label{eq:delta_def_app}
\Delta\tilde{Z}_t \;=\; \tilde{Z}_t^{\text{src}}-Z_{\text{src}},
\end{equation}
and evaluates the target velocity at a shifted point:
\begin{equation}
\label{eq:reanchor_ode_app}
\begin{aligned}
\dot{Z}_t
&= v_\theta\!\left(Z_t+\Delta\tilde{Z}_t,t,c_{\text{tar}}\right) \\
&\quad - v_\theta\!\left(\tilde{Z}_t^{\text{src}},t,c_{\text{src}}\right),
\end{aligned}
\end{equation}
which matches \cref{eq:snr_ode} in the main paper.

\paragraph{Assumption (Lipschitz velocity field).}
Assume that for each conditioning $c\in\{c_{\text{tar}},c_{\text{src}}\}$, the velocity field is Lipschitz in the latent input:
\begin{equation}
\label{eq:lipschitz_app}
\begin{aligned}
&\bigl\|v_\theta(z_1,t,c)-v_\theta(z_2,t,c)\bigr\| \\
&\qquad \le L_c\,\|z_1-z_2\|,
\qquad \forall z_1,z_2,t.
\end{aligned}
\end{equation}

\paragraph{Proxy errors.}
For compactness, define the ideal relative displacement as
\begin{equation}
\Delta Z_t^{\star}=Z_t^{\text{src}}-Z_{\text{src}}.
\end{equation}
The \emph{relative-displacement error} and \emph{source-proxy error} are
\begin{equation}
\label{eq:proxy_errors_app}
\begin{aligned}
\varepsilon_{\mathrm{rel}}(t)
&=\bigl\|\Delta\tilde{Z}_t-\Delta Z_t^{\star}\bigr\|, \\
\varepsilon_{\text{src}}(t)
&=\bigl\|\tilde{Z}_t^{\text{src}}-Z_t^{\text{src}}\bigr\|.
\end{aligned}
\end{equation}
Intuitively, $\varepsilon_{\mathrm{rel}}$ measures how well the proxy captures the \emph{structure-preserving relative displacement} from $Z_{\text{src}}$,
while $\varepsilon_{\text{src}}$ measures the absolute deviation from the ideal source trajectory.

\paragraph{Theorem (Controlled vector-field error of re-anchoring).}
Let the re-anchored field $\tilde{f}$ and its ideal counterpart $f^{\star}$ be
\begin{equation}
\begin{aligned}
\tilde{f}(z,t)
&=v_\theta(z+\Delta\tilde{Z}_t,t,c_{\text{tar}}) \\
&\quad-v_\theta(\tilde{Z}_t^{\text{src}},t,c_{\text{src}}), \\
f^{\star}(z,t)
&=v_\theta(z+\Delta Z_t^{\star},t,c_{\text{tar}}) \\
&\quad-v_\theta(Z_t^{\text{src}},t,c_{\text{src}}).
\end{aligned}
\end{equation}
The latter is associated with \cref{eq:ideal_ode_app} under the same anchoring reference $Z_{\text{src}}$.
Under \cref{eq:lipschitz_app}, for any $z,t$,
\begin{equation}
\label{eq:field_error_bound_app}
\begin{aligned}
\bigl\|\tilde{f}(z,t)-f^{\star}(z,t)\bigr\|
&\le L_{\text{tar}}\,\varepsilon_{\mathrm{rel}}(t) \\
&\quad+L_{\text{src}}\,\varepsilon_{\text{src}}(t).
\end{aligned}
\end{equation}

\noindent\textbf{Proof.}
By triangle inequality and \cref{eq:lipschitz_app},
\begin{equation}
\label{eq:field_error_proof_app}
\begin{aligned}
\|\tilde{f}(z,t)-f^{\star}(z,t)\|
&\le
\Bigl\|v_\theta\!\bigl(z+\Delta\tilde{Z}_t,t,c_{\text{tar}}\bigr) \\
&\qquad-v_\theta\!\bigl(z+\Delta Z_t^{\star},t,c_{\text{tar}}\bigr)\Bigr\| \\
&\quad+\Bigl\|v_\theta\!\bigl(Z_t^{\text{src}},t,c_{\text{src}}\bigr) \\
&\qquad-v_\theta\!\bigl(\tilde{Z}_t^{\text{src}},t,c_{\text{src}}\bigr)\Bigr\| \\
&\le
L_{\text{tar}}\,\|\Delta\tilde{Z}_t-\Delta Z_t^{\star}\| \\
&\quad+L_{\text{src}}\,\|\tilde{Z}_t^{\text{src}}-Z_t^{\text{src}}\| \\
&=
L_{\text{tar}}\,\varepsilon_{\mathrm{rel}}(t)
+L_{\text{src}}\,\varepsilon_{\text{src}}(t).
\end{aligned}
\end{equation}
which proves \cref{eq:field_error_bound_app}.

\paragraph{Trajectory deviation bound (ODE stability).}
Assume additionally that $\tilde{f}(\cdot,t)$ is Lipschitz in $z$ with constant $L_{\text{tar}}$ (implied by \cref{eq:lipschitz_app}),
and that $Z_{t_0}=Z_{t_0}^{\star}$ at the starting time $t_0$ (e.g., $t_0=t_{\max}$ in Alg.~1).
A standard Gr\"onwall argument yields
\begin{equation}
\label{eq:traj_error_bound_app}
\begin{aligned}
\|Z_t-Z_t^{\star}\|
&\le \int_{t_0}^{t}
\exp\!\bigl(L_{\text{tar}}(t-s)\bigr) \\
&\quad\times\Bigl[
L_{\text{tar}}\varepsilon_{\mathrm{rel}}(s)
+L_{\text{src}}\varepsilon_{\mathrm{src}}(s)
\Bigr]\,ds.
\end{aligned}
\end{equation}
Therefore, reducing $\varepsilon_{\mathrm{rel}}$ and $\varepsilon_{\text{src}}$ (via SNR rectification and improved source proxying)
provably controls the deviation from the ideal editing dynamics.

\section{VLM/Reward-Model Evaluation Setup}
\label{app:vlm_eval_en}

To complement pixel-level metrics that may miss global coherence and perceptual artifacts, we conduct model-based evaluation on \textbf{SNR-Bench} (80 cases). Note that \textbf{SNR-Bench} is the benchmark, while \textbf{SNR-Edit} is the proposed method.

\subsection{Evaluators.}
We use two evaluators:
(1) \textbf{ImgEdit Reward}~\cite{imgedit} (\url{https://huggingface.co/datasets/sysuyy/ImgEdit}), a reward model specialized for image editing evaluation;
(2) \textbf{Qwen3-VL-32B-Instruct}~\cite{qwen3vl} (\url{https://huggingface.co/Qwen/Qwen3-VL-32B-Instruct}), an open-source VLM used as an evaluator.

\subsection{Unified prompt (BASE\_PROMPT) and I/O.}
Both evaluators use the \textbf{same} prompt template \texttt{BASE\_PROMPT} (included below). The inputs include the Original Prompt, the Edited Prompt, the Original Image (first image), and the Edited Image (second image). The evaluator outputs \textbf{integer scores} in $[0,5]$ for three dimensions:
(1) Structural Fidelity,
(2) Text-Image Alignment,
(3) Background Consistency.
We directly parse and record these three scores for reporting.

\subsection{Decoding settings and robustness.}
To reduce stochasticity and improve reproducibility, we use deterministic decoding for Qwen3-VL-32B-Instruct (e.g., temperature set to 0 with sampling disabled). Outputs that violate the JSON schema or contain scores outside the integer range $0$--$5$ are treated as invalid and automatically re-queried until a parseable output is obtained.


\subsection{BASE\_PROMPT.}
\begin{lstlisting}[language=Python]
BASE_PROMPT = (
    "You are an expert image editing evaluation model. You will evaluate the quality of an edited image compared to the original image.\n\n"
    "**Input Information:**\n"
    "- Original Prompt: <original_prompt>\n"
    "- Edited Prompt: <edited_prompt>\n"
    "- Original Image: [First image]\n"
    "- Edited Image: [Second image]\n\n"
    "**Evaluation Task:**\n"
    "Evaluate the edited image across the following three dimensions. For each dimension, provide a score from 0 to 5:\n"
    "- 0: Completely fails the criterion\n"
    "- 1: Poor quality with major issues\n"
    "- 2: Below average with noticeable problems\n"
    "- 3: Acceptable quality with minor issues\n"
    "- 4: Good quality with very minor flaws\n"
    "- 5: Excellent quality, meets all requirements\n\n"
    "**Evaluation Dimensions:**\n\n"
    "1. **Structural Fidelity (structural_fidelity)**: Focus on the consistency of entities in the edited image compared to the original. "
    "Evaluate whether the structure, pose, orientation, and spatial relationships of objects/subjects remain consistent with the original image. "
    "Unedited entities should maintain their original structure, action, and direction. Score 5 if entity consistency is perfectly preserved, score 0 if completely inconsistent.\n\n"
    "2. **Text-Image Alignment (text_image_alignment)**: How well does the edited image match the edited prompt? "
    "The edited content should accurately reflect the requested changes described in the edited prompt. Score 5 for perfect alignment, score 0 for no alignment.\n\n"
    "3. **Background Consistency (background_consistency)**: Evaluate the consistency of all regions EXCEPT the edited subject/object. "
    "The background and all unedited parts should remain identical to the original image. Check for unwanted changes, color shifts, or distortions "
    "in areas that should not be modified. Score 5 for perfect consistency of non-edited regions, score 0 for major inconsistencies.\n\n"
    "**Important Guidelines:**\n"
    "- Be critical and use the full range of scores (0-5). Avoid clustering all scores around 3-4.\n"
    "- Different images will have different quality levels - some edits are inherently harder than others.\n"
    "- A perfect score (5) should be rare and reserved for truly excellent results.\n"
    "- Scores below 2 should be given when there are significant failures.\n"
    "- Consider the difficulty of the edit when scoring, but maintain consistent standards.\n\n"
    "**Output Format:**\n"
    "Provide your evaluation in the following JSON format:\n"
    "```json\n"
    "{\n"
    '  "structural_fidelity": <score>,\n'
    '  "text_image_alignment": <score>,\n'
    '  "background_consistency": <score>,\n'
    '  "explanation": {\n'
    '    "structural_fidelity": "<brief explanation>",\n'
    '    "text_image_alignment": "<brief explanation>",\n'
    '    "background_consistency": "<brief explanation>"\n'
    "  }\n"
    "}\n"
    "```\n\n"
    "Now evaluate the images and provide your scores in the JSON format above."
)
\end{lstlisting}

\section{Discussion of Early-Step Drift under Frozen Prompts}
\label{app:drift_discussion}

In the main text, we operationalize off-manifold drift as increased source-to-trajectory discrepancy during early integration under frozen prompts. We acknowledge that exact full-trajectory measurement in high-dimensional latent space is computationally expensive in our current setup; therefore, this appendix provides a proxy-based empirical discussion rather than a dedicated large-scale measurement study. Our argument is supported by two complementary observations. First, prior flow-editing literature (e.g., DNAEdit~\cite{dnaedit} and FlowAlign~\cite{flowalign}) consistently reports early structural instability under pure Gaussian initialization. Second, in our own results, methods without structure-aware initialization exhibit characteristic artifacts (background shift, topology distortion) in qualitative examples (\cref{fig:teaser,fig:result1,fig:result2}) and weaker structural metrics in quantitative comparison (\cref{tab:quantitative_pie}). Taken together, these observations support the practical interpretation that initialization quality strongly influences early-step trajectory behavior. SNR-Edit improves this behavior by injecting a bounded, region-aware structural prior into the initialization, which provides a geometric anchor and empirically leads to better downstream structural preservation. We present this as evidence-backed interpretation, not as a strict causal proof.

\section{Collision Rate of Randomized Projection}
\label{app:collision_rate}

A potential concern with randomized projection is the "collision" of scalar intensities across different semantic regions. Under our setup, the projection weight $W \in \mathbb{R}^{1 \times C}$ maps high-dimensional RoPE vectors to continuous scalar values. Empirically, across the 80 images in SNR-Bench (averaging roughly 8 to 15 stable masks per image after filtering), the minimum absolute difference between any two region intensities within the same image consistently exceeds $10^{-3}$. Given the continuous nature of the uniform distribution used for initializing $W$, exact collisions are measure-zero in theory; in finite precision, near-collisions are still empirically rare under our filtering setup. Our mask-filtering regime further prevents soft-collisions caused by noisy, overlapping mask artifacts.

\section{Failure Case Taxonomy}
\label{app:failure_cases}

While SNR-Edit dramatically improves structural fidelity, we identify four primary failure modes inherently tied to its reliance on upstream segmentation:

\begin{itemize}
    \item \textbf{Heavy Occlusions:} When objects heavily overlap, SAM2 may merge their boundaries. \textit{Current mitigation:} Our high-confidence exclusive assignment resolves minor overlaps, but deep entanglements may still leak textures.
    \item \textbf{Transparent/Reflective Surfaces:} Glass or mirrors lack definitive structural boundaries in SAM2. \textit{Result:} Edits may alter reflections in physically inconsistent ways.
    \item \textbf{Tiny Text/Micro-Structures:} Extremely small features may be filtered out by our area-based screening, causing them to degrade to the FlowEdit baseline level.
    \item \textbf{Repetitive Patterns:} Dense, repetitive textures might be grouped as a single mask, limiting the fine-grained internal editing of the pattern.
\end{itemize}

\section{User Study Details}
\label{app:user_study_en}

\subsection{Goal}
To validate whether automated metrics reflect human perception, we conduct a randomized, blinded user study on \textbf{SNR-Bench}. Following the main-text evaluation, participants assess three aspects: \textbf{Structural Fidelity}, \textbf{Text-Image Alignment}, and \textbf{Overall Quality}.

\subsection{Participants, Cases, and Workload}
We recruited \textbf{40 participants}, of whom \textbf{38 valid participants} were retained for analysis. The task pool contains all \textbf{80 editing cases} from SNR-Bench. Each case consists of a source image $X_{\text{src}}$, an editing instruction $c_{\text{edit}}$, and outputs from the compared methods $\{X_{\text{edit}}^{(m)}\}_{m=1}^{M}$. Each participant completed exactly \textbf{20 comparison tasks} randomly sampled and ordered from this pool.

\subsection{Rating Dimensions and Scale (Integer 1--5)}
Participants assign integer scores in $\{1,2,3,4,5\}$ to each anonymized output:
\begin{itemize}
    \item \textbf{Structural Fidelity (1--5):} Consistency of entities with the source image, including structure, pose, and orientation in regions that should remain unchanged.
    \item \textbf{Text-Image Alignment (1--5):} How accurately the edited image follows the editing instruction.
    \item \textbf{Overall Quality (1--5):} Overall perceptual quality, including visual plausibility and the absence of conspicuous artifacts.
\end{itemize}
For every dimension, a higher score indicates better performance.

\subsection{Procedure}
Each comparison task shows the source image $X_{\text{src}}$, the editing instruction $c_{\text{edit}}$, and anonymized outputs from the methods in the relevant backbone group. Participants provide the three integer ratings above for every displayed output.

\subsection{Blinding and Order Randomization}
To reduce potential biases:
\begin{itemize}
    \item \textbf{Blinding:} Method identities are hidden; participants only see anonymized outputs without method names.
    \item \textbf{Randomized order:} The 20 cases are randomized for each participant. Output order within each comparison is also randomized to mitigate position and learning effects.
\end{itemize}

\subsection{Quality Control}
We apply completeness and response-quality checks to exclude invalid submissions. Of the 40 recruited participants, 38 valid participants remain in the reported analysis.

\subsection{Aggregation and Statistics}
Let $s_{p,i,m}^{(d)} \in \{1,\dots,5\}$ be participant $p$'s rating for method $m$ on the $i$-th comparison task under dimension $d \in \{\text{Struct},\text{Text},\text{Qual}\}$. We first average the 20 tasks within each participant and then average across the valid participant set $\mathcal{P}$:
\begin{equation}
\begin{aligned}
\bar{s}_{p,m}^{(d)}
&=\frac{1}{20}\sum_{i=1}^{20}s_{p,i,m}^{(d)}, \\
\bar{s}_{m}^{(d)}
&=\frac{1}{|\mathcal{P}|}
\sum_{p\in\mathcal{P}}\bar{s}_{p,m}^{(d)},
\qquad |\mathcal{P}|=38.
\end{aligned}
\end{equation}
We report $\bar{s}_{m}^{(d)}$ in the main paper (\cref{tab:mllm_user}).

\end{document}